\documentclass[11pt,a4paper]{article}
\usepackage[colorlinks=true,linkcolor=blue]{hyperref}

\usepackage{cmap}
\usepackage[utf8x]{inputenc}
\usepackage[T1]{fontenc}
\usepackage[english]{babel}
\usepackage{amssymb,amsmath}
\usepackage{amsthm}
\usepackage{bbm}
\usepackage{microtype}
\usepackage{lmodern}

{
\rmfamily
\DeclareFontShape{T1}{lmr}{m}{sc}{<->ssub*cmr/m/sc}{}
\DeclareFontShape{T1}{lmr}{b}{sc}{<->ssub*cmr/b/sc}{}
\DeclareFontShape{T1}{lmr}{bx}{sc}{<->ssub*cmr/bx/sc}{}
}

\newcommand{\thmheadercommand}[1]{\textbf{\scshape{}#1.\\*}}

\newtheoremstyle{yannthm}{\topsep}{\topsep}{\slshape}{}{\scshape\bfseries}{.}{.5em}{%
\thmname{#1}\thmnumber{ #2}\thmnote{#3}%
}
\newtheoremstyle{yannthm2}{\topsep}{\topsep}{}{}{\scshape\bfseries}{.}{.5em}{%
\thmname{#1}\thmnumber{ #2}\thmnote{#3}%
}

\def\d{\operatorname{d}\!{}}

\def\R{{\mathbb{R}}}

\renewcommand{\geq}{\geqslant}

\newcommand{\deq}{\mathrel{\mathop{:}}=}

\def\eps{\varepsilon}
\renewcommand{\epsilon}{\varepsilon}
\renewcommand{\phi}{\varphi}

\def\ds{\displaystyle}

\DeclareMathOperator{\Cov}{Cov}
\let\oldPr\Pr
\renewcommand{\Pr}{\oldPr\nolimits}
\newcommand{\E}{\mathbb{E}}

\DeclareMathOperator{\Id}{Id}

\DeclareMathOperator{\diag}{diag}

\newenvironment{dem}[1][]{\begin{proof}[\thmheadercommand{Proof#1}]~\newline\ignorespaces}{\end{proof}}

{
\theoremstyle{yannthm}
\newtheorem{defi}{Definition}
\newtheorem*{defi*}{Definition}
\newtheorem{prop}[defi]{Proposition}
\newtheorem*{prop*}{Proposition}
\newtheorem{thm}[defi]{Theorem}
\newtheorem*{thm*}{Theorem}
\newtheorem{lem}[defi]{Lemma}
\newtheorem*{lem*}{Lemma}
\newtheorem{cor}[defi]{Corollary}
\newtheorem*{cor*}{Corollary}

\newtheorem*{ex*}{Example}

\newtheorem*{subenonce}{}

\theoremstyle{yannthm2}

\newtheorem*{exo*}{Exercise}

\newtheorem*{rem*}{Remark}

\newtheorem*{subenonce2}{}

}

\newcommand{\transp}[1]{#1^{\!\top}\!}

\title{Online Natural Gradient as a Kalman Filter}

\author{Yann Ollivier}
\date{}

\newcommand{\gaussian}{\mathcal{N}}

\newcommand{\Ptheta}{P^\theta}
\newcommand{\Py}{P^{\hat y}}
\newcommand{\Pthetay}{P^{\theta\hat y}}
\newcommand{\deltatheta}{\delta\hspace{-.05em}\theta}
\newcommand{\transf}{\Phi}

\numberwithin{equation}{section}

\begin{document}

\maketitle

\begin{abstract}
We cast Amari's natural gradient in statistical learning as a specific
case of Kalman filtering.
Namely, applying an extended Kalman filter to estimate a fixed unknown
parameter of a probabilistic model from a series of observations, is
rigorously equivalent to estimating this parameter via
an online stochastic natural gradient descent on the log-likelihood of the
observations.

In the i.i.d.\ case, this relation
is a consequence of the ``information filter'' phrasing of
the extended Kalman filter.  
In the recurrent (state space, non-i.i.d.) case, we prove that the joint Kalman filter over states
and parameters is a natural gradient on top of real-time recurrent
learning (RTRL), a classical algorithm to train recurrent models.

This exact algebraic correspondence provides relevant interpretations for natural
gradient hyperparameters such as learning rates or initialization and
regularization of the Fisher information matrix.
\end{abstract}

In statistical learning, stochastic gradient descent is a widely used
tool to estimate the parameters of a model from empirical data,
especially when the parameter dimension and the amount of data are large
\cite{bottoulecun2003} (such as is typically the case with neural 
networks, for instance).
The \emph{natural gradient} \cite{Amari1998} is a tool from information
geometry, which aims at correcting several shortcomings of the widely
ordinary stochastic gradient descent, such as
its sensitivity to rescalings or simple changes of
variables in parameter space \cite{gradnn}. The natural gradient modifies the ordinary
gradient by using the information geometry of the statistical model, via
the Fisher information matrix (see formal definition in
Section~\ref{sec:natgrad}; see also \cite{martensnatgrad}). The natural gradient comes with a theoretical guarantee of
asymptotic optimality \cite{Amari1998} that the ordinary gradient lacks, and with the theoretical knowledge
and various connections from
information geometry, e.g., \cite{Amari2000book,igo}. In large dimension, its
computational complexity makes approximations
necessary, e.g.,
\cite{TONGA,gradnn,riemaNN,grosse2015scaling,martens2015optimizing};
this has limited its adoption despite many desirable
theoretical properties.

The \emph{extended Kalman filter} (see e.g., the textbooks
\cite{simon2006kalmanbook,sarkka_book,jazwinski_book}) is a generic and
effective tool to estimate in real
time the state of a nonlinear dynamical system, from noisy measurements
of some part or some function of the system.  (The ordinary Kalman filter
deals with \emph{linear} systems.) Its use in navigation systems (GPS,
vehicle control, spacecraft...), time series analysis, econometrics,
etc.\ \cite{sarkka_book},
is extensive  to the point it can been described as ``one of the great
discoveries of mathematical engineering'' \cite{kalmanmatlab}.

The goal of this text is to show that the natural gradient, when applied
online, is a particular case of the extended Kalman filter.  Indeed, the
extended Kalman filter can be used to estimate the parameters of a
statistical model (probability distribution), by viewing the parameters
as the hidden state of a ``static'' dynamical system, and viewing i.i.d.~samples
as noisy observations depending on the parameters \footnote{For this we
slightly extend the definition of the Kalman filter to include discrete
observations, by defining (Def.~\ref{def:kalman}) the measurement error as $T(y)-\hat y$
instead of $y-\hat y$, where $T$ is the sufficient statistics of an
exponential family model for output noise with mean $\hat y$. This reduces to the standard filter for Gaussian
output noise, and naturally covers categorical outputs as often used in
statistical learning (with $\hat y$ the class probabilities in a softmax
classifier and $T$ a ``one-hot'' encoding of $y$).}. We show that doing so
is exactly equivalent to performing an online stochastic natural gradient
descent (Theorem~\ref{thm:natkal}).

This results in a rigorous dictionary between the natural gradient objects
from statistical learning, and the objects appearing in Kalman filtering;
for instance, a larger learning rate for the natural gradient descent
exactly corresponds to a fading memory in the Kalman filter
(Proposition~\ref{prop:rates}).

Table~\ref{tab:corr} lists a few correspondences between objects from
the Kalman filter side and from the natural gradient side, as results
from the theorems and propositions below. Note that the correspondence is
one-sided: the online natural gradient is exactly an extended Kalman filter, but
only corresponds to a particular use of the Kalman filter for parameter
estimation problems (i.e., with static dynamics on the parameter part of the
system).

Beyond the static case, we also consider the learning of the parameters
of a general dynamical system, where subsequent observations exhibit
temporal patterns instead of being i.i.d.; in statistical learning this is called a
\emph{recurrent} model, for instance, a recurrent neural network.  We
refer to \cite{Jaeger_tutorial} for an introduction to recurrent models
in statistical learning (recurrent neural networks) and the afferent
techniques (including Kalman filters), and to \cite{Haykin_book} for a
clear, in-depth treatment of Kalman filtering for recurrent models.  We
prove (Theorem~\ref{thm:rtrlkal}) that the extended Kalman filter applied
jointly to the state and parameter, amounts to a natural gradient
on top of \emph{real-time recurrent learning} (RTRL), a classical (and
costly) online algorithm for recurrent network training
\cite{Jaeger_tutorial}.

\begin{table}
\small
\begin{tabular}{ll}
\hline
\multicolumn{2}{l}{\emph{iid (static, non-recurrent) model $\hat y_t=h(\theta,u_t)$}}\\
Extended Kalman filter on static & Online natural
gradient on $\theta$ with\\parameter $\theta$& learning rate $\eta_t=1/(t+1)$
\\
Covariance matrix $P_t$  & Fisher information matrix $J_t=\eta_t
P_t^{-1}$
\\
Bayesian prior $P_0$ & Fisher matrix initialization $J_0=P_0^{-1}$
\\
Fading memory & Larger or constant learning rate
\\
Fading memory+constant prior & Fisher matrix regularization
\\
\hline
\multicolumn{2}{l}{\emph{Recurrent (state space) model $\hat y_t=\Phi(\hat
y_{t-1},\theta,u_t)$}}\\
Extended Kalman filter on $(\theta,\hat y)$ & RTRL+natural gradient+state
correction\\
Covariance of $\theta$ alone, $\Ptheta$ & Fisher matrix $J_t=\eta_t \,(\Ptheta)^{-1}$\\
Correlation between $\theta$ and $\hat y_t$ & RTRL gradient estimate
$\partial \hat y_t/\partial \theta$
\\\hline
\end{tabular}
\caption{Kalman filter objects vs natural gradient objects. The inputs
are $u_t$, the predicted values are $\hat y_t$, and the model parameters
are $\theta$.}
\label{tab:corr}
\end{table}

Thus, we provide a bridge between techniques from large-scale statistical
learning (natural gradient, RTRL) and a central object from mathematical
engineering, signal processing, and estimation theory.  Casting the
natural gradient as a specific case of the extended Kalman filter is an
instance of the provocative statement from \cite{ljung83} that ``there is
only one recursive identification method'' that is optimal on quadratic
functions. Indeed, the online natural gradient descent fits into the
framework of \cite[\S3.4.5]{ljung83}. Arguably, this statement is limited
to linear models, and for non-linear models one would expect different
algorithms to coincide only at a certain order, or asymptotically;
however, all the correspondences presented below are exact.

\paragraph*{Related work.}
In the i.i.d.\ (static) case, the natural gradient/Kalman filter
correspondence follows from the information filter phrasing of Kalman
filtering \cite[\S6.2]{simon2006kalmanbook} by relatively direct
manipulations. Nevertheless, we could find no reference in the literature
explicitly identifying the two.  \cite{singhalwu1988} is an
early example of the use of Kalman filtering for training feedforward
neural networks in statistical learning, but does not mention the natural
gradient. \cite{ruck1992comparative} argue that for neural networks,
backpropagation, i.e., ordinary gradient descent, ``is a degenerate form of the
extended Kalman filter''.
\cite{bertsekas96}
identifies the extended Kalman filter with a Gauss--Newton gradient
descent for the specific case of nonlinear regression.
\cite{freitas2000hierarchical}
interprets process noise in the static Kalman filter as
an adaptive, per-parameter
learning rate, thus akin to a preconditioning matrix.
\cite{simandl2001CramerRao} uses the Fisher information matrix to study
the variance of parameter estimation in Kalman-like filters, without
using a natural gradient; \cite{bottoulecun2003} comment on the similarity between
Kalman filtering and a version of Amari's natural gradient for the specific case of least squares
regression; \cite{martensnatgrad} and \cite{gradnn} mention the
relationship between natural gradient and the Gauss--Newton Hessian
approximation; \cite{patel_kalmansgd} exploits the
relationship between second-order gradient descent and Kalman filtering
in specific cases including linear regression; \cite{li2017information} use a natural
gradient descent over Gaussian distributions for an auxiliary problem arising in Kalman-like Bayesian
filtering, a problem independent from the one treated here.

For the recurrent
(non-i.i.d.) case, our result is that joint Kalman filtering is
essentially a natural gradient on top of the classical RTRL algorithm for
recurrent models \cite{Jaeger_tutorial}. 
\cite{williams1992training} already observed that starting with the Kalman
filter and introducing drastic
simplifications (doing away with the covariance matrix) results in RTRL,
while
\cite[\S5]{Haykin_book} contains statements that can be
interpreted as relating Kalman filtering and preconditioned RTRL-like
gradient descent for recurrent models (Section \ref{sec:rec_statement}).

\paragraph*{Perspectives.} In this text our goal is to derive the precise
correspondence between natural gradient and Kalman filtering for
parameter estimation (Thm.~\ref{thm:natkal}, Prop.~\ref{prop:rates},
Prop.~\ref{prop:regul}, Thm.~\ref{thm:rtrlkal}), and to work
out an exact dictionary between the mathematical objects on both sides.
This correspondence suggests several possible
venues for research, which nevertheless are not explored here.

First, the correspondence with the Kalman filter brings new
interpretations and suggestions for several natural gradient
hyperparameters, such as Fisher matrix initialization, equality between
Fisher matrix decay rate and learning rate, or amount of regularization
to the Fisher matrix (Section~\ref{sec:statement}). The natural gradient
can be quite sensitive to these hyperparameters. A first step would be to
test the matrix decay rate and regularization values suggested by the
Bayesian interpretation (Prop.~\ref{prop:regul}) and see if they help
with the natural gradient, or if these suggestions are overriden by the
various approximations needed to apply the natural gradient in practice.
These empirical tests are beyond the scope of the present study.

Next, since statistical learning deals with either continuous or
categorical data, we had to extend the usual Kalman filter to such a
setting. Traditionally, non-Gaussian
output models have been treated by applying a nonlinearity to a standard
Gaussian noise (Section~\ref{sec:proofs}). Instead, modeling the measurement noise as an exponential family
(Appendix and Def.~\ref{def:kalman}) allows for a unified treatment of
the standard case (Gaussian output noise with known variance), of
discrete categorical observations, or other exponential noise models
(e.g., Gaussian noise with unknown variance). We did not test the
empirical consequences of this choice, but it certainly makes the
mathematical treatment flow smoothly, in particular the view of the
Kalman filter as preconditioned gradient descent
(Prop.~\ref{prop:Kalmanasgrad}).

Neither the natural gradient nor the extended Kalman filter scale well to
large-dimensional models as currently used in machine learning, so that
approximations are required.  The correspondence raises the possibility
that various methods developed for Kalman filtering (e.g., particle or
unscented filters) or for natural gradient approximations (e.g., matrix
factorizations such as the Kronecker product \cite{martens2015optimizing}
or quasi-diagonal reductions \cite{gradnn,riemaNN}) could be transferred from one viewpoint to the other.

In statistical learning, other means have been developed to attain the
same asymptotic efficiency as the natural gradient, notably trajectory
averaging (e.g.~\cite{polyakjuditsky1992}, or \cite{martensnatgrad} for
the relationship to natural gradient) at little algorithmic cost. One may
wonder if these can be generalized to filtering problems.

Proof techniques could be transferred as well: for instance, Amari \cite{Amari1998}
gave a strong but sometimes informal argument that the natural gradient
is Fisher-efficient, i.e., the resulting parameter estimate is asymptotically optimal for the Cramér--Rao
bound; alternate proofs could be obtained by transferring related
statements for the extended Kalman filter, e.g., combining techniques
from \cite{simandl2001CramerRao,boutayeb1997EKFconv,ljung83}.

\paragraph*{Organization of the text.} In Section~\ref{sec:setting} we
set the notation, recall the definition of the natural gradient
(Def.~\ref{def:natgrad}), and explain how Kalman filtering can be used for
parameter estimation in statistical learning (Section~\ref{sec:kalman}); the definition of the
Kalman filter is included in Def.~\ref{def:kalman}.
Section~\ref{sec:static} gives the main statements for viewing the
natural gradient as an instance of an extended Kalman filter for i.i.d.\ 
observations (static systems), first intuitively via a heuristic
asymptotic argument (Section~\ref{sec:heuristics}), then rigorously
(Thm.~\ref{thm:natkal}, Prop.~\ref{prop:rates}, Prop.~\ref{prop:regul}).
The proof of these results appears in Section~\ref{sec:proofs} and sheds
some light on the geometry of Kalman filtering. Finally, the case of
non-i.i.d.\ observations (recurrent or state space model) is treated in
Section~\ref{sec:rec}.

\paragraph*{Acknowledgments.} Many thanks to Silvère Bonnabel, Gaétan
Marceau-Caron, and the anonymous reviewers for their careful reading of
the manuscript, corrections, and suggestions for the presentation and
organization of the text.  I would also like to thank Shun-ichi Amari,
Frédéric Barbaresco, and Nando de Freitas for additional comments and for
pointing out relevant references.

\section{Problem setting, natural gradient, Kalman filter}
\label{sec:setting}

\subsection{Problem setting}
In statistical learning,
we have a series of observation pairs
$(u_1,y_1),\ldots,(u_t,y_t),\ldots$ and want to predict $y_t$ from $u_t$
using a probabilistic model $p_\theta$. Assume for now that $y_t$ is
real-valued (regression problem) and that the model for
$y_t$ is a Gaussian centered on a predicted value $\hat y_t$, with known
covariance matrix $R_t$, namely
\begin{equation}
y_t=\hat y_t+\gaussian(0,R_t) ,
\qquad
\hat y_t=h(\theta,u_t)
\end{equation}
The function $h$ may represent any computation, for instance, a feedforward neural network with input $u$, parameters
$\theta$, and output $\hat y$. The goal is to find the parameters $\theta$ such that the prediction
$\hat y_t=h(\theta,u_t)$
is as close as possible to $y_t$: the loss function is
\begin{equation}
\ell_t=\frac12 \transp{(\hat y_t-y_t)}R_t^{-1}(\hat y_t-y_t)=-\ln p(y_t|\hat y_t)
\end{equation}
up to an additive constant.

For non-Gaussian outputs, we assume that the noise model
on
$y_t$ given $\hat y_t$ belongs to an exponential family,
namely, that $\hat y_t$ is the
mean parameter of an exponential family of distributions
\footnote{
The Appendix contains a reminder on exponential families. An \emph{exponential family of probability distributions} on $y$, with sufficient statistics
$T_1(y),\ldots,T_K(y)$, and with parameter $\beta\in \R^K$, is given by
\begin{equation}
p_\beta(y)\deq\frac{1}{Z(\beta)}\,\mathrm{e}^{\sum_k \beta_k
T_k(y)}\,\lambda(\d y)
\end{equation}
where $Z(\beta)$ is a normalizing constant, and $\lambda(\d y)$ is any
reference
measure on $y$. For instance, if $y\in \R^K$, $T_k(y)=y_k$ and $\lambda(\d y)$
is a Gaussian measure centered on $0$, by varying $\beta$ one gets all
Gaussian measures with the same covariance matrix and another mean. $y$ may be discrete,
e.g., 
Bernoulli distributions correspond to $\lambda$
the uniform measure on $y\in \{0,1\}$ and a single sufficient
statistic $T(0)=0$, $T(1)=1$. Often, the \emph{mean parameter} $\bar
T\deq\E_{y\sim p_\beta} T(y)$ is a more convenient parameterization than
$\beta$.  Exponential families maximize entropy (minimize information
divergence from $\lambda$) for a given
mean of $T$.}
over $y_t$; we again define the
loss function as $\ell_t\deq -\ln p(y_t|\hat y_t)$, and the output
noise $R_t$ can be defined 
as the covariance matrix of the sufficient statistics of $y_t$ given this
mean (Def.~\ref{def:kalman}). For a Gaussian output noise this works as expected. For instance, for a classification problem, the output is
categorical, $y_t\in\{1,\ldots, K\}$, and $\hat y_t$ will be the set of
probabilities $\hat y_t=(p_1,\ldots,p_{K-1})$ to have
$y_t=1,\ldots,{K-1}$. In that case $R_t$ is the $(K-1)\times (K-1)$ matrix
$(R_t)_{kk'}=\diag(p_k)-p_k p_{k'}$. (The last probability $p_K$ is
determined by the others via $\sum p_k=1$ and has to be excluded to
obtain a non-degenerate parameterization and an invertible covariance
matrix $R_t$.)

This convention allows us to
extend the definition of the Kalman filter to such a
setting (Def.~\ref{def:kalman}) in a natural way, just by replacing the measurement error
$y_t-\hat y_t$ with $T(y_t)-\hat y_t$, with $T$ the sufficient statistics
for the exponential family. (For Gaussian noise this is the same, as
$T(y)$ is $y$.)

In neural network terms, this means that the \emph{output layer} of the
network is fed to a loss function that is the log-loss of an exponential
family, but places no restriction on the rest of the model.

%


\paragraph*{General notation.} In
statistical learning, the external inputs or regressor variables are often
denoted $x$
. In Kalman filtering, $x$ often denotes the state of
the system, while the external inputs are often $u$. Thus 
we will avoid $x$ altogether and denote by $u$ the inputs and
by $s$ the state of
the system.

The variable to be predicted at time $t$ will be $y_t$, and $\hat y_t$ is
the corresponding prediction. In general $\hat y_t$ and $y_t$ may be
different objects in that $\hat y_t$ encodes a full probabilistic
prediction for $y_t$. For Gaussians with known variance, $\hat y_t$ is
just the predicted mean of $y_t$, so in this case $y_t$ and $\hat y_t$
are the same type of object.  For Gaussians with unknown variance, $\hat
y$ encodes both the mean and second moment of $y$. For discrete
categorical data, $\hat y$ encodes the probability of each possible
outcome $y$.

Thus, the formal setting for this text is as follows: we are given a
sequence of finite-dimensional observations $(y_t)$ with each $y_t\in
\R^{\dim(y)}$, a sequence of inputs $(u_t)$ with each $u_t\in
\R^{\dim(u)}$, a parametric model $\hat y=h(\theta,u_t)$ with parameter
$\theta\in \R^{\dim(\theta)}$ and $h$ some fixed smooth function from
$\R^{\dim(\theta)}\times \R^{\dim(u)}$ to $\R^{\dim(\hat y)}$. We are
given an exponential family (output noise model) $p(y|\hat y)$ on $y$
with mean parameter $\hat y$ and sufficient statistics $T(y)$ (see the
Appendix), and we define the loss function $\ell_t\deq -\ln p(y_t|\hat y_t)$.

The natural gradient descent on parameter $\theta_t$ will use the Fisher
matrix $J_t$. The Kalman filter will have posterior covariance matrix
$P_t$.

For multidimensional quantities $x$ and $y=f(x)$, we denote by $\frac{\partial
y}{\partial x}$ the Jacobian matrix of $y$ w.r.t.\ $x$, whose $(i,j)$
entry is $\frac{\partial f_i(x)}{\partial x_j}$. This satisfies the chain
rule $\frac{\partial z}{\partial y}\frac{\partial y}{\partial
x}=\frac{\partial
z}{\partial x}$. With this convention, 
gradients of real-valued functions are \emph{row} vectors, so that a
gradient descent takes the form $x\gets x - \eta\, \transp{(\partial
f/\partial x)}$.

For a column vector $u$, $u^{\otimes 2}$ is synonymous with
$u\transp{u}$, and with $\transp{u}u$ for a row vector.


\subsection{Natural gradient descent}
\label{sec:natgrad}

A standard approach to optimize the
parameter $\theta$ of a probabilistic model, given a sequence of
observations $(y_t)$, is an online gradient descent
\begin{equation}
\theta_t\gets \theta_{t-1} - \eta_t \transp{\frac{\partial \ell_t(y_t)}{\partial
\theta}}
\end{equation}
with learning rate $\eta_t$.  This simple gradient descent is
particularly suitable for large datasets and large-dimensional models
\cite{bottoulecun2003}, but has several
practical and theoretical shortcomings. For instance, it uses the same
non-adaptive learning rate for all parameter components. Moreover, simple
changes in parameter encoding or in data
presentation (e.g., encoding black and white in images by 0/1 or 1/0) can
result in different learning performance.

This motivated the introduction of the
\emph{natural gradient} \cite{Amari1998}. 
 It is built to achieve
invariance with respect to parameter re-encoding; in particular,
learning become insensitive to the characteristic scale of each parameter
direction, so that different directions naturally get suitable learning
rates. The natural gradient is the only general way to achieve
such invariance \cite[\S2.4]{Amari2000book}.

The natural gradient preconditions
the gradient descent with
$J(\theta)^{-1}$ where $J$ is the \emph{Fisher information matrix} 
\cite{Kullback} with
respect to the parameter $\theta$. For a smooth probabilistic model
$p(y|\theta)$ over a random variable $y$ with parameter $\theta$, the
latter
is defined as
\begin{equation}
J(\theta)\deq \E_{y\sim p(y|\theta)} \left[{\frac{\partial \ln
p(y|\theta)}{\partial \theta}}^{\otimes 2}\right]=-\E_{y\sim
p(y|\theta)}\left[
\frac{\partial^2\ln p(y|\theta)}{\partial \theta^2}\right]
\end{equation}
Definition~\ref{def:natgrad} below formally
introduces the \emph{online} natural gradient.
If the model for $y$ involves an input $u$, then an expectation or
empirical average over the input is introduced in the definition of $J$
\cite[\S8.2]{Amari2000book} \cite[\S5]{martensnatgrad}.

However, this comes at a large computational cost for large-dimensional
models: just storing the Fisher matrix already costs $O((\dim
\theta)^2)$. Various strategies are available to approximate the natural
gradient for complex models such as neural networks, using diagonal or
block-diagonal approximation schemes for the Fisher matrix, e.g.,
\cite{TONGA,gradnn,riemaNN,grosse2015scaling,martens2015optimizing}.

\begin{defi}[ (Online natural gradient)]
\label{def:natgrad}
Consider a statistical model with parameter $\theta$ that predicts an
output $y$ given an input $u$. Suppose that the prediction takes the form
$y\sim p(y|\hat y)$ where $\hat y=h(\theta,u)$ depends on the input via a
model $h$ with parameter $\theta$.
Given observation pairs $(u_t,y_t)$, the goal is to minimize, online, the loss function
\begin{equation}
\sum_t \ell_t(y_t), \qquad \ell_t(y)\deq -\ln p(y|\hat y_t)
\end{equation}
as a function of $\theta$.

The \emph{online natural gradient} maintains a current estimate
$\theta_t$ of the parameter $\theta$, and a current approximation $J_t$ of the
Fisher matrix. The parameter is estimated by a gradient descent with
preconditioning matrix $J_t^{-1}$, namely
\begin{align}
\label{eq:natgradJ}
J_t &\gets (1-\gamma_t) J_{t-1} + \gamma_t \,\E_{y\sim 
p(y|\hat y_t)} \left[\frac{\partial \ell_t(y)}{\partial
\theta}^{\otimes 2}\right]
\\
\label{eq:natgradtheta}
\theta_t &\gets \theta_{t-1} -\eta_t \, J_t^{-1}\transp{\left(
\frac{\partial \ell_t(y_t)}{\partial \theta}
\right)}
\end{align}
with learning rate $\eta_t$ and Fisher matrix decay rate $\gamma_t$.
\end{defi}

In the Fisher matrix update, the expectation over all possible values
$y\sim p(y|\hat y)$ can often be computed algebraically, but this is
sometimes computationally bothersome (for instance, in neural networks,
it requires $\dim(\hat y_t)$ distinct backpropagation steps \cite{gradnn}). A
common solution \cite{APF00,TONGA,gradnn,PB13} is to just use the value $y=y_t$ (\emph{outer
product} approximation) instead of the expectation over $y$. Another is to use a
Monte Carlo approximation with a single sample of $y\sim p(y|\hat y_t)$
\cite{gradnn,riemaNN}, namely, using the gradient of a synthetic sample
instead of the actual observation $y_t$ in the Fisher matrix. These
latter two solutions are often confused; only the latter provides an
unbiased estimate, see discussion in \cite{gradnn,PB13}.

The online ``smoothed'' update of the Fisher matrix in
\eqref{eq:natgradJ} mixes past and present estimates (this or similar
updates are used in \cite{TONGA,riemaNN}). The reason is at least
twofold. First, the ``genuine'' Fisher matrix involves an
expectation over the inputs $u_t$ \cite[\S8.2]{Amari2000book}: this can be approximated online only
via a moving average over inputs (e.g., $\gamma_t=1/t$ realizes an
equal-weight average over all inputs seen so far).
Second, the expectation over $y\sim p(y|\hat y_t)$ in \eqref{eq:natgradJ}
is often replaced with a Monte Carlo estimation with only one value of
$y$, and averaging over time compensates for this Monte Carlo sampling.

As a consequence, since $\theta_t$ changes over time, this means that the
estimate $J_t$ mixes values obtained at different values of $\theta$, and
converges to the Fisher matrix only if $\theta_t$ changes slowly, i.e.,
if $\eta_t\to 0$.  The correspondence below with Kalman filtering
suggests using $\gamma_t=\eta_t$.

\subsection{Kalman filtering for parameter estimation}
\label{sec:kalman}

One possible definition of the extended Kalman filter is as follows
\cite[\S15.1]{simon2006kalmanbook}. We
are trying to estimate the current state of a dynamical system $s_t$
whose evolution equation is known but whose precise value is unknown; at
each time step, we have access to a noisy measurement $y_t$ of a quantity
$\hat y_t=h(s_t)$ which depends on this state.

The Kalman filter
maintains an approximation of a Bayesian posterior on $s_t$ given the
observations $y_1,\ldots,y_t$. The posterior distribution after
$t$ observations is approximated by a Gaussian with mean $s_t$ and
covariance matrix $P_t$. (Indeed, Bayesian posteriors always tend to
Gaussians asymptotically under mild conditions, by the Bernstein--von
Mises theorem \cite{van2000asymptotic}.) The Kalman filter
prescribes a way to update $s_t$ and $P_t$ when new observations become
available.

The Kalman filter update is summarized in Definition~\ref{def:kalman}
below.  It is built to provide the \emph{exact} value of the Bayesian
posterior in the case of \emph{linear} dynamical systems with Gaussian
measurements and a Gaussian prior. In that sense, it is exact at first
order.

The Kalman filtering viewpoint on a statistical learning problem is that
we are facing a system with hidden variable $\theta$, with an unknown
value that does not evolve in time, and that the observations $y_t$ bring
more and more information on $\theta$.  Thus, a statistical learning
problem can be tackled by applying the extended Kalman filter to the
unknown variable $s_t=\theta$, whose underlying dynamics from time $t$ to
time $t+1$ is just to remain unchanged ($f=\Id$ and noise on $s$ is $0$
in Definition~\ref{def:kalman}). In such a setting, the posterior
covariance matrix $P_t$ will generally tend to $0$ as observations
accumulate and the parameter is identified better\footnote{But $P_t$ must
still be maintained even if it tends to $0$, since it is used to update
the parameter at the correct rate.} (this occurs at rate $1/t$ for the
basic filter, which estimates from all $t$ past observations at time $t$,
or at other rates if fading memory is included, see below). The
initialization $\theta_0$ and its covariance $P_0$ can be interpreted as
Bayesian priors on $\theta$ \cite{singhalwu1988,ljung83}.

We will refer to this as a \emph{static} Kalman filter.  In the static
case and without fading memory, the posterior covariance $P_t$ after $t$
observations will decrease like $O(1/t)$, so that the parameter gets
updated by $O(1/t)$ after each new observation. Introducing fading memory
for past observations (equivalent to adding noise on $\theta$ at each
step, $Q_t\propto P_{t|t-1}$ in Def.~\ref{def:kalman}) leads to a larger
covariance and faster updates.

\paragraph*{An example: Feedforward neural networks.} The Kalman approach
above can be applied to any parametric statistical model. For instance
\cite{singhalwu1988} treat the case of a feedforward neural network. In
our setting this is described as follows. Let $u$ be the input of the
model and $y$ the true (desired) output. A feedforward neural network can be
described as a function $\hat y=h(\theta,u)$ where $\theta$ is the set of
all parameters of the network, where $h$ represents all computations
performed by the network on input $u$, and $\hat y$ encodes the network
prediction for the value of the output $y$ on input $u$. For categorical
observations $y$, $\hat y$ is usually a set of predicted probabilities
for all possible classes; while for regression problems, $\hat y$ is
directly the predicted value. In both cases, the error function to be
minimized can be defined as $\ell(y)\deq -\ln p(y|\hat y)$: in the
regression case, $\hat y$ is interpreted as a mean of a Gaussian model on
$y$, so that $-\ln p(y|\hat y)$ is the square error up to a constant.

Training the neural network amounts to estimating the network parameter
$\theta$ from the observations. Applying a static Kalman filter for this
problem \cite{singhalwu1988} amounts to using Def.~\ref{def:kalman} with
$s=\theta$, $f=\Id$ and $Q=0$. At first glance this looks quite different
from the common gradient descent (backpropagation) approach for neural
networks. The
backpropagation operation is represented in the Kalman filter by the
computation of $H=\frac{\partial h(s,u)}{\partial s}$
\eqref{eq:defH} where $s$ is the parameter. We show that the additional
operations of the Kalman filter correspond to using a natural gradient
instead of a vanilla gradient.

Unfortunately, for models with high-dimensional parameters such as neural
networks, the Kalman filter is computationally costly and requires
block-diagonal approximations for $P_t$ (which is a square matrix of size
$\dim \theta$); moreover, computing $H_t=\partial \hat y_t/\partial
\theta$ is needed in the filter, and requires doing one separate
backpropagation for each component of the output $\hat y_t$.

\section{Natural gradient as a Kalman filter: the static (i.i.d.) case}
\label{sec:static}

We now write the explicit correspondence between an online natural
gradient to estimate the parameter of a statistical model from i.i.d.\
observations, and a static extended Kalman filter. We first give a
heuristic argument that outlines the main ideas from the proof
(Section~\ref{sec:heuristics}).

Then we state the formal correspondences. First, the static Kalman filter
corresponds to an online natural gradient with learning rate $1/t$
(Thm.~\ref{thm:natkal}). The rate $1/t$ arises because such a filter
takes into account all previous evidence without decay factors (and with
process noise $Q=0$ in the Kalman filter), thus the posterior covariance
matrix decreases like $O(1/t)$. Asymptotically, this is the
optimal rate in statistical learning \cite{Amari1998}.
(Note, however, that the online natural gradient and extended Kalman
filter are identical at every time step, not only asymptotically.)

The $1/t$ rate is often too slow in practical applications,
especially when starting far away from an optimal parameter value. The
natural gradient/Kalman filter correspondence is not specific to the
$O(1/t)$ rate. Larger learning rates in the natural gradient correspond
to a \emph{fading memory} Kalman filter (adding process noise $Q$
proportional to the posterior covariance at each step, corresponding to a
decay factor for the weight of previous observations); this is
Proposition~\ref{prop:rates}. In such a setting, the posterior covariance
matrix in the Kalman filter does not decrease like $O(1/t)$; for
instance, a fixed decay factor for the fading memory corresponds to a
constant learning rate.

Finally, a fading memory in the Kalman filter may erase prior Bayesian
information $(\theta_0,P_0)$ too fast; maintaining the weight of the
prior in a fading memory Kalman filter is treated in
Proposition~\ref{prop:regul} and corresponds, on the natural gradient
side, to a so-called weight decay \cite{Bishop_book} towards $\theta_0$ together with a regularization
of the Fisher matrix, at specific rates.

\subsection{Natural gradient as a Kalman filter: heuristics}
\label{sec:heuristics}

As a first ingredient in the correspondence, we interpret Kalman filters as gradient descents:
the extended Kalman filter actually performs a gradient descent on the
log-likelihood of each new observation, with preconditioning matrix equal to
the posterior covariance matrix. This is Proposition~\ref{prop:Kalmanasgrad}
below. This relies on having an exponential family as the output noise
model.

Meanwhile, the natural gradient uses the Fisher matrix as a
preconditioning matrix. The Fisher matrix is the average Hessian
of log-likelihood, thanks to the classical double definition of the
Fisher matrix as square gradient or Hessian,
$J(\theta)=\E_{y\sim p(y|\theta)} \left[\frac{\partial \ln p(y)}{\partial \theta}^{\otimes
2}\right]=-\E_{y\sim p(y|\theta)} \left[\frac{\partial^2 \ln p(y)}{\partial
\theta^2}\right]$ for any probabilistic model $p(y|\theta)$
\cite{Kullback}.

Assume that the probability of the data given the parameter $\theta$ is
approximately Gaussian,
$p(y_1,\ldots,y_t|\theta)\propto\exp(-\transp{(\theta-\theta^\ast)}\Sigma^{-1}(\theta-\theta^\ast))$
with covariance $\Sigma$. This often holds asymptotically thanks to the
Bernstein--von Mises theorem; moreover, the posterior covariance $\Sigma$
typically decreases like $1/t$. Then the Hessian (w.r.t.\ $\theta$) of the total
log-likelihood of $(y_1,\ldots,y_t)$ is $\Sigma^{-1}$, the inverse
covariance of $\theta$. So the \emph{average} Hessian per data point, the
Fisher matrix $J$,
is approximately $J\approx \Sigma^{-1}/t$. Since a Kalman filter to estimate
$\theta$ is essentially a gradient descent preconditioned with
$\Sigma$, it will be the same as using a natural gradient with learning
rate $1/t$. Using a fading memory Kalman filter will estimate $\Sigma$
from fewer past observations and provide larger learning rates.

%

Another way to understand the link between natural gradient and Kalman
filter is as a second-order Taylor expansion of data log-likelihood.
Assume that the total data log-likelihood at time $t$, $L_t(\theta)\deq-
\sum_{s=1}^t \ln p(y_s|\theta)$, is approximately quadratic as a function
of $\theta$, with a minimum at $\theta^\ast_t$ and a Hessian $h_t$,
namely, $L_t(\theta)\approx \frac12
\transp{(\theta-\theta^\ast_t)}h_t(\theta-\theta^\ast_t)$. Then when new
data points become available, this quadratic approximation would be updated
as follows (online Newton method):
\begin{align}
\label{eq:onlineNewt1}
h_t&\approx h_{t-1} + \partial_\theta^2 (-\ln
p(y_t|\theta^\ast_{t-1}))
\\
\label{eq:onlineNewt2}
\theta^\ast_t &\approx \theta^\ast_{t-1} - h_t^{-1} \,\partial_\theta (-\ln
p(y_t|\theta^\ast_{t-1}))
\end{align}
and indeed these
are \emph{equalities} for a quadratic log-likelihood. Namely, the update
of $\theta^\ast_t$ is a gradient ascent on log-likelihood,
preconditioned by the inverse Hessian (Newton method). Note that $h_t$
grows like $t$ (each data point adds its own contribution). Thus,
$h_t$ is $t$ times the empirical average of the Hessian, i.e., approximately
$t$ times the Fisher matrix of the model ($h_t\approx t J$). So this update is approximately
a natural gradient descent with learning rate $1/t$.

Meanwhile, the Bayesian posterior on $\theta$ (with uniform prior) after
observations $y_1,\ldots,y_t$ is proportional to $e^{-L_t}$ by definition
of $L_t$. If $L_t\approx \frac12
\transp{(\theta-\theta^\ast_t)}h_t(\theta-\theta^\ast_t)$, this is a Gaussian distribution centered
at $\theta^\ast_t$ with covariance matrix $h_t^{-1}$. The Kalman filter
is built to
maintain an approximation $P_t$ of this covariance matrix $h_t^{-1}$, and
then performs a gradient step preconditioned on $P_t$ similar to
\eqref{eq:onlineNewt2}.

The simplest situation corresponds to an asymptotic rate $O(1/t)$, i.e.,
estimating the parameter based on all past evidence; the update
\eqref{eq:onlineNewt1} of the Hessian is additive, so that $h_t$ grows
like $t$ and $h_t^{-1}$
in \eqref{eq:onlineNewt2}
produces an effective learning rate $O(1/t)$.
Introducing a decay factor for older observations, multiplying the term
$h_{t-1}$ in \eqref{eq:onlineNewt1}, produces a fading memory effect and
results in larger learning rates.

\bigskip

These heuristics justify the statement from \cite{ljung83} that ``there is
only one recursive identification method''. Close to an optimum (so that
the Hessian is positive), all second-order algorithms are essentially an
online Newton step \eqref{eq:onlineNewt1}-\eqref{eq:onlineNewt2}
approximated in various ways.

But even though this heuristic argument appears to be approximate or asymptotic, the
correspondence between online natural gradient and Kalman filter
presented below is exact at every time step.

\subsection{Statement of the correspondence, static (i.i.d.) case}
\label{sec:statement}

For the statement of the correspondence, we assume
that the output noise on $y$
given $\hat y$ is modelled by an exponential family with mean parameter
$\hat y$. This covers the traditional Gaussian case
$y=\gaussian(\hat y,\Sigma)$ with fixed $\Sigma$ often used in Kalman
filters. The Appendix contains necessary background on exponential
families.

%

\begin{thm}[ (Natural gradient as a static Kalman filter)]
\label{thm:natkal}
These two algorithms are identical under the correspondence
$(\theta_t,J_t)\leftrightarrow (s_t,P_t^{-1}/(t+1))$:
\begin{enumerate}
\item The online natural gradient (Def.~\ref{def:natgrad}) with learning
rates $\eta_t=\gamma_t=1/(t+1)$, applied to learn the parameter $\theta$
of a model that
predicts observations $(y_t)$ with inputs $(u_t)$, using a
probabilistic model $y\sim p (y|\hat y)$ with $\hat y=h(\theta,u)$, where
$h$ is any model and $p(y|\hat y)$ is an
exponential family with mean parameter $\hat y$.
\item The extended Kalman filter (Def.~\ref{def:kalman}) to estimate the state $s$ from
observations $(y_t)$ and inputs $(u_t)$,
using a
probabilistic model $y\sim p (y|\hat y)$ with $\hat y=h(s,u)$ and
$p(y|\hat y)$ an
exponential family with mean parameter $\hat y$, 
with static dynamics and no
added noise on $s$
($f(s,u)=s$ and $Q=0$ in Def.~\ref{def:kalman}).
\end{enumerate}

Namely, if at startup $(\theta_0,J_0)=(s_0,P_0^{-1})$, then
$(\theta_t,J_t)=(s_t,P_t^{-1}/(t+1))$ for all $t\geq 0$.
\end{thm}

The correspondence is exact only if the Fisher metric is updated before
the parameter in the natural gradient descent (as in
Definition~\ref{def:natgrad}).

The
correspondence with a Kalman filter provides an interpretation for
various hyper-parameters of online natural gradient descent. 
%
In particular, $J_0=P_0^{-1}$ can be interpreted as the inverse
covariance of a Bayesian prior on $\theta$ \cite{singhalwu1988}. This
relates the initialization $J_0$ of the Fisher matrix to the
initialization of $\theta$: for instance, in neural networks it is
recommended to initialize the weights according to a Gaussian of
covariance $\diag(1/\text{fan-in})$ (number of incoming weights) for each
neuron; interpreting this as a Bayesian prior on weights, one may
recommend to initialize the Fisher matrix to the inverse of this
covariance, namely,
\begin{equation}J_0\gets\diag(\text{fan-in})\end{equation} Indeed this seemed to
perform quite well in small-scale experiments.

\paragraph*{Learning rates, fading memory, and metric decay rate.}
Theorem~\ref{thm:natkal} exhibits a $1/(t+1)$ learning rate for the
online natural gradient. This is because the static Kalman filter for
i.i.d.\ observations approximates
the maximum a posteriori (MAP) of the parameter $\theta$ based on all
past observations; MAP and
maximum likelihood estimators change by $O(1/t)$ when a new data point is
observed.

However, for nonlinear systems, optimality of the $1/t$ rate only occurs
asymptotically, close enough to the optimum.  In general, a $1/(t+1)$
learning rate is far from optimal if optimization does not start close to
the optimum or if one is not using the exact Fisher matrix $J_t$ or
covariance matrix $P_t$.

Larger effective learning rates are achieved thanks to so-called ``fading
memory'' variants of the Kalman filter, which put less weight on older
observations. For instance, one may multiply the log-likelihood of
previous points by a forgetting factor $(1-\lambda_t)$ before each new
observation. This is equivalent to an additional step $P_{t-1}\gets
P_{t-1}/(1-\lambda_t)$ in the Kalman filter, or to the addition of an
artificial process noise $Q_t$ proportional to $P_{t-1}$ in the model. Such
strategies are reported to often improve performance,
especially when the data do not truly follow the model
\cite[\S5.5, \S7.4]{simon2006kalmanbook}, \cite[\S5.2.2]{Haykin_book}.
See for instance \cite{bertsekas96} for the relationship between Kalman fading
memory and gradient descent learning rates (in a particular case).

\begin{prop}[ (Natural gradient rates and fading memory)]
\label{prop:rates}
Under the same model and assumptions as in Theorem~\ref{thm:natkal},
the following two algorithms are identical via the correspondence
$(\theta_t,J_t)\leftrightarrow (s_t,\eta_t P_t^{-1})$:
\begin{itemize}
\item An online natural gradient step with learning rate $\eta_t$ and
metric decay rate $\gamma_t$

\item A fading memory Kalman filter with an additional step $P_{t-1} \gets
P_{t-1}/(1-\lambda_t)$ before the transition step; such a filter iteratively optimizes a weighted
log-likelihood function $L_t$ of recent observations, with decay
$(1-\lambda_t)$ at each step, namely:
\begin{equation}
L_t(\theta)=\ln p_\theta(y_t)+(1-\lambda_t)\,L_{t-1}(\theta)
\,,\qquad L_0(\theta)\deq -\frac12 
\transp{(\theta-\theta_0)}P_0^{-1}(\theta-\theta_0)
\end{equation}
\end{itemize}
provided the following relations are satisfied:
\begin{align}
&\eta_t=\gamma_t,\qquad P_0=\eta_0 J_0^{-1},
\\&
\label{eq:rates}
1-\lambda_t=
\eta_{t-1}/\eta_t - \eta_{t-1} \qquad \text{for $t \geq 1$}
\end{align}
\end{prop}

For example, taking $\eta_t=1/(t+\mathrm{cst})$ corresponds to
$\lambda_t=0$, no decay for older observations, and an initial covariance
$P_0=J_0^{-1}/\mathrm{cst}$. Taking a constant learning rate
$\eta_t=\eta_0$ corresponds to a constant decay factor $\lambda=\eta_0$.

The proposition above computes the fading memory decay factors
$1-\lambda_t$ from the natural gradient learning rates $\eta_t$ via
\eqref{eq:rates}.
In the other direction, one can start with the decay factors $\lambda_t$ and obtain
the learning rates $\eta_t$ via the cumulated sum of weights $S_t$: $S_0\deq
1/\eta_0$ then $S_t\deq
(1-\lambda_t)S_{t-1}+1$, then $\eta_t\deq1/S_t$.
This clarifies how $\lambda_t=0$ corresponds to
$\eta_t=1/(t+\mathrm{cst})$ where the constant is $S_0$.

The learning rates also control the weight given to the Bayesian prior
and to the starting point $\theta_0$. For instance, with
$\eta_t=1/(t+t_0)$ and large $t_0$, the gradient descent will move away slowly from
$\theta_0$; in the Kalman interpretation this
corresponds to $\lambda_t=0$ and a small initial covariance
$P_0=J_0^{-1}/t_0$ around $\theta_0$, so that the prior weighs
as much as $t_0$ observations.

This result suggests to set $\gamma_t=\eta_t$ in the online natural
gradient descent of Definition~\ref{def:natgrad}. The intuitive
explanation for this setting is as follows: Both the Kalman filter and
the natural gradient build a second-order approximation of the
log-likelihood of past observations as a function of the parameter
$\theta$, as explained in Section~\ref{sec:heuristics}. Using a fading
memory corresponds to putting smaller weights on past observations; these
weights affect the first-order and the second-order parts of the
approximation in the same way. In the gradient viewpoint, the learning
rate $\eta_t$ corresponds to the first-order term (comparing
\eqref{eq:natgradtheta} and \eqref{eq:onlineNewt2}) while the Fisher matrix
decay rate corresponds to the rate at which the second-order information
is updated. Thus, the setting $\eta_t=\gamma_t$ in
the natural gradient corresponds to using the same decay weights for the
first-order and second-order expansion of the log-likelihood of past
observations.

Still, one should keep in mind that the extended Kalman filter is itself
only an approximation for nonlinear systems. Moreover, from a statistical
point of view, the second-order object $J_t$ is higher-dimensional than
the first-order information, so that estimating $J_t$ based on more past
observations may be more stable. Finally, for large-dimensional problems
the Fisher matrix is always approximated, which affects optimality of the
learning rates. So in practice, considering
$\gamma_t$ and $\eta_t$ as hyperparameters to be tuned independently may
still be beneficial, though $\gamma_t=\eta_t$ seems a good place to
start.


\paragraph*{Regularization of the Fisher matrix and Bayesian priors.} A
potential downside of fading memory in the Kalman filter is that
the Bayesian interpretation is partially lost,
because the Bayesian prior is forgotten too
quickly. For instance, with a constant learning rate, the weight of the
Bayesian prior decreases exponentially; likewise, with $\eta_t=O(1/\sqrt{t})$, the filter
essentially works with the $O(\sqrt{t})$ most recent observations, while
the weight of the prior decreases like $\approx
e^{-\sqrt{t}}$ (as does the weight of the earliest observations; this is
the product $\prod (1-\lambda_t)$).  But
precisely,
when working with fewer data points one may wish the prior to
play a greater role.

The Bayesian interpretation can be restored by explicitly optimizing a
combination of the log-likelihood of recent points, and the
log-likelihood of the prior. This is implemented in
Proposition~\ref{prop:regul}.

From the natural gradient viewpoint, this translates both as a
regularization of the Fisher matrix (often useful in practice to
numerically stabilize its inversion) and of the gradient step. With a
Gaussian prior $\gaussian(\theta_{\mathrm{prior}},\Id)$, this manifests
as an additional step towards $\theta_\mathrm{prior}$ and adding
$\eps.\Id$ to the Fisher matrix, known respectively as weight decay and
Tikhonov regularization \cite[\S3.3, \S5.5]{Bishop_book} in statistical
learning.

\begin{prop}[ (Bayesian regularization of the Fisher matrix)]
\label{prop:regul}
Let $\pi=\gaussian(\theta_{\mathrm{prior}},\Sigma_0)$ be a Gaussian prior on
$\theta$.
Under the same model and assumptions as in Theorem~\ref{thm:natkal},
the following two algorithms are equivalent:
\begin{itemize}
\item A modified fading memory Kalman filter that iteratively optimizes
$L_t(\theta)+n_\mathrm{prior}\ln
\pi(\theta)$ where $L_t$ is a weighted
log-likelihood function of recent observations with decay $(1-\lambda_t)$:
\begin{equation}
L_t(\theta)=\ln p_\theta(y_t)+(1-\lambda_t)\, L_{t-1}(\theta), \qquad
L_0\deq 0
\end{equation}
initialized with $P_0=\frac{\eta_1}{1+n_\mathrm{prior}\eta_1}\Sigma_0$.

\item A regularized online natural gradient step with learning rate $\eta_t$ and
metric decay rate $\gamma_t$, initialized with $J_0=\Sigma_0^{-1}$,
\begin{align}
\theta_t &\gets \theta_{t-1} -\eta_t \, \left(J_t+\eta_t n_\mathrm{prior} \Sigma_0^{-1}\right)^{-1}\left(
\transp{\frac{\partial \ell_t(y_t)}{\partial \theta}}
+\lambda_t n_\mathrm{prior} \Sigma_0^{-1}(\theta-\theta_{\mathrm{prior}})
\right)
\end{align}
\end{itemize}
provided the following relations are satisfied:
\begin{equation}
\eta_t=\gamma_t,\qquad 
1-\lambda_t=
\eta_{t-1}/\eta_t - \eta_{t-1},\qquad \eta_0\deq \eta_1
\end{equation}
\end{prop}


Thus, the regularization terms are fully determined by choosing
the learning rates $\eta_t$, a prior such as
$\gaussian(0,1/\text{fan-in})$ (for neural networks), and a value of
$n_\mathrm{prior}$ such as $n_\mathrm{prior}=1$ (the prior weighs as much as $n_\mathrm{prior}$ data points). This holds 
both for regularization of the Fisher matrix $J_t+\eta_t n_\mathrm{prior}
\Sigma_0^{-1}$, and for regularization of the parameter via the extra
gradient step $\lambda_t
n_\mathrm{prior} \Sigma_0^{-1}(\theta-\theta_\text{prior})$.


The relative strength of regularization in the Fisher matrix
decreases like $\eta_t$. In particular, a constant learning rate results
in a constant regularization.

The added gradient step $\lambda_t n_\mathrm{prior}
\Sigma_0^{-1}(\theta-\theta_\text{prior})$ is modulated by $\lambda_t$
which depends on $\eta_t$; this extra term pulls towards the prior
$\theta_\mathrm{prior}$. The Bayesian viewpoint guarantees that this
extra term will not ultimately prevent convergence of the gradient
descent (as the influence of the prior vanishes when the number of
observations increases).

\bigskip

It is not clear how much these recommendations for natural gradient
descent coming from its Bayesian interpretation are sensitive to
using only an approximation of the Fisher matrix.

\subsection{Proofs for the static case}
\label{sec:proofs}

The proof of Theorem~\ref{thm:natkal} starts with the
interpretation of the Kalman filter as a gradient descent
(Proposition~\ref{prop:Kalmanasgrad}).

We first recall the exact definition and the notation we use for the
extended Kalman filter.

\begin{defi}[ (Extended Kalman filter)]
\label{def:kalman}
Consider a dynamical system
with state $s_{t}$, inputs $u_{t}$ and outputs $y_{t}$,
\begin{equation}
s_{t}=f(s_{t-1},u_{t})+\gaussian(0,Q_{t}),
\qquad
\hat y_{t}=h(s_{t},u_{t}),\qquad
y_{t}\sim p(y|\hat y_{t})
\end{equation}
where $p(\cdot|\hat y)$ denotes an exponential family with mean parameter $\hat y$
(e.g., $y=\gaussian(\hat y,R)$ with fixed covariance matrix $R$).

The \emph{extended Kalman filter} for this dynamical system
estimates the current state $s_t$ given observations $y_1,\ldots,y_t$ in
a Bayesian fashion.
At each time, the Bayesian posterior distribution of the state given
$y_1,\ldots,y_t$
is approximated by a Gaussian $\gaussian(s_t,P_t)$ so that $s_t$ is the
approximate maximum a posteriori, and $P_t$ is the approximate posterior
covariance matrix. (The prior is $\gaussian(s_0,P_0)$ at time $0$.)
Each time a new observation $y_{t}$ is available, these estimates are
updated as follows.

The transition step (before
observing $y_{t}$) is
\begin{align}
s_{t|{t-1}} & \gets f(s_{t-1},u_{t})
\\
\label{eq:KFdefF}
F_{t-1} & \gets \left.\frac{\partial f}{\partial s}\right|_{(s_{t-1},u_{t})}
\\
\label{eq:transP}
P_{t|{t-1}} & \gets F_{t-1} P_{t-1} \transp{F_{t-1}} + Q_{t}
\\
\hat y_{t} &\gets h(s_{t|{t-1}},u_{t})
\end{align}
and the observation step after observing $y_{t}$ is
\begin{align}
E_{t} &\gets \text{sufficient statistics}(y_{t})-\hat y_{t}
\\
R_{t} &\gets \Cov (\text{sufficient statistics}(y)|\hat y_{t})
\label{eq:KFdefR}
\intertext{(these are just the error $E_{t}
= y_{t}-\hat y_{t}$ and the covariance matrix $R_t=R$
for a
Gaussian model $y=\gaussian(\hat y,R)$ with known $R$)}
H_t &\gets \left.\frac{\partial h}{\partial
s}\right|_{(s_{t|{t-1}},u_{t})} \label{eq:defH}
\\
K_{t} &\gets P_{t|{t-1}}\transp{H_{t}}\left(H_{t} P_{t|{t-1}}
\transp{H_{t}}+R_{t}\right)^{-1}\label{eq:KFdefK}
\\
P_{t} &\gets \left(\Id-K_{t}H_{t}\right)P_{t|{t-1}} \label{eq:KFPupdate}
\\
s_{t} &\gets s_{t|{t-1}} + K_{t}E_{t}\label{eq:KFsupdate}
\end{align}
\end{defi}

For non-Gaussian output noise, the definition of $E_t$ and $R_t$ above
via the mean parameter $\hat y$ of an exponential family,
differs from the practice of modelling non-Gaussian noise via a nonlinear
function applied to Gaussian noise. This allows for a straightforward
treatment of various output models, such as discrete outputs or Gaussians
with unknown variance. In the
Gaussian case with known variance our definition is fully standard.
\footnote{Non-Gaussian output noise is often modelled in
Kalman filtering via a continuous nonlinear function applied to a
Gaussian noise \cite[13.1]{simon2006kalmanbook}; this cannot easily represent discrete random variables.
Moreover, since the filter linearizes the function around the $0$ value
of the noise \cite[13.1]{simon2006kalmanbook}, the noise is still
implicitly Gaussian, though with a state-dependent
variance.\label{ft:nongaussian}}

\bigskip

The proof starts with the interpretation of the Kalman filter as a
gradient descent preconditioned by $P_t$. Compare this result and
Lemma~\ref{lem:infofilter} to
\cite[(5.68)--(5.73)]{Haykin_book}.

\begin{prop}[ (Kalman filter as preconditioned gradient descent)]
\label{prop:Kalmanasgrad}
The update of the state $s$ in a Kalman filter can be seen as an online gradient
descent on data log-likelihood, with preconditioning matrix $P_t$. More
precisely, denoting $\ell_t(y)\deq -\ln p(y|\hat y_t)$, the update~\eqref{eq:KFsupdate} is equivalent to
\begin{equation}
s_t=s_{t|t-1}-P_t \transp{\left(\frac{\partial \ell_t(y_t)}{\partial s_{t|t-1}}\right)}
\end{equation}
where in the derivative, $\ell_t$ depends on $s_{t|t-1}$ via $\hat
y_t=h(s_{t|t-1},u_t)$.
\end{prop}

\begin{lem}[ (Errors and gradients)]
\label{lem:errisgrad}
When the output model is an exponential family with mean parameter $\hat
y_t$,
the error $E_t$ is related to the gradient of the log-likelihood of the
observation $y_t$ with respect to the prediction $\hat y_t$ by
\[
E_{t} = R_{t} \transp{\left(\frac{\partial \ln 
p(y_{t}|\hat y_{t})}{\partial \hat y_t}\right)}
\]
\end{lem}

\begin{dem}[ of the lemma]
For a Gaussian $y_t=\gaussian(\hat y_t,R)$, this is just a direct
computation. For a general exponential family,
consider the natural parameter $\beta$ of the exponential family which
defines the law of $y$, namely, $p(y|\beta)=\exp(\sum_i \beta_i T_i(y))/Z(\beta)$
with sufficient statistics $T_i$ and normalizing constant $Z$. An
elementary computation (Appendix, \eqref{eq:expder}) shows that
\begin{equation}
\frac{\partial \ln p(y|\beta)}{\partial \beta_i}=T_i(y)-\E T_i=T_i(y)-\hat y_i
\end{equation}
by definition of the mean parameter $\hat y$. Thus,
\begin{equation}
E_t=T(y_t)-\hat y_t =\transp{\left(\frac{\partial \ln p(y_t|\beta)}{\partial \beta}\right)}
\end{equation}
where the derivative is with respect to the natural parameter $\beta$. To
express the derivative with respect to $\hat y$, we apply the chain rule
\[
\frac{\partial \ln p(y_t|\beta)}{\partial \beta}=\frac{\partial \ln
p(y_t|\hat y)}{\partial \hat y}\frac{\partial \hat y}{\partial \beta}
\]
and use the fact that, for exponential families, the Jacobian matrix of the
mean parameter $\frac{\partial \hat y}{\partial \beta}$ is equal to the
covariance matrix $R_t$ of the
sufficient statistics (Appendix, \eqref{eq:expjac} and
\eqref{eq:J=cov_comp}).
\end{dem}

\begin{lem}
\label{lem:KR=PH}
The extended Kalman filter satisfies $K_tR_t=P_t \transp{H_t}$.
\end{lem}

\begin{dem}[ of the lemma]
This relation is known, e.g., \cite[(6.34)]{simon2006kalmanbook}.
Indeed,
using the definition of $K_t$, we have
$K_tR_t=K_t(R_t+H_tP_{t|t-1}\transp{H_t})-K_tH_tP_{t|t-1}\transp{H_t}=P_{t|t-1}\transp{H_t}-K_t
H_tP_{t|t-1}\transp{H_t}=(\Id-K_tH_t)P_{t|t-1}\transp{H_t}=P_t\transp{H_t}$.
\end{dem}

\begin{dem}[ of Proposition~\ref{prop:Kalmanasgrad}]
By definition of the Kalman filter we have
$s_{t}=s_{t|{t-1}}+K_{t}E_{t}$. By Lemma~\ref{lem:errisgrad},
$\ds E_{t}=R_{t}\transp{\left(\frac{\partial \ell_t}{\partial \hat y_t}\right)}$.
Thanks to Lemma~\ref{lem:KR=PH} we find
$\ds s_{t}=s_{t|{t-1}}+K_{t}R_{t}\transp{\left(\frac{\partial \ell_t}{\partial
\hat y_t}\right)}=s_{t|{t-1}}+P_{t}\transp{H_{t}}\transp{\left(\frac{\partial
\ell_t}{\partial \hat y_t}\right)}=s_{t|{t-1}}+P_{t}\transp{\left(\frac{\partial
\ell_t}{\partial \hat y_t} H_{t}\right)}$. But by the definition of $H$, $H_{t}$
is $\ds \frac{\partial \hat y_{t}}{\partial s_{t|{t-1}}}$ so that $\ds \frac{\partial
\ell_t}{\partial \hat y_t} H_{t}$ is $\ds \frac{\partial \ell_t}{\partial s_{t|{t-1}}}$.
\end{dem}

The first part of the next lemma is known as the information filter in the Kalman filter
literature, and states that the observation step for $P$ is additive when
considered on $P^{-1}$ \cite[\S6.2]{simon2006kalmanbook}: after each observation, the Fisher information matrix of
the latest observation is added to $P^{-1}$.

\begin{lem}[ (Information filter)]
\label{lem:infofilter}
The update \eqref{eq:KFdefK}--\eqref{eq:KFPupdate} of $P_t$ in the
extended Kalman filter is equivalent to
\begin{equation}
P_t^{-1} \gets P_{t|t-1}^{-1} + \transp{H_t} R_t^{-1} H_t
\end{equation}
(assuming $P_{t|t-1}$ and $R_t$ are invertible).

In particular, for
static dynamical systems ($f(s,u)=s$ and $Q_t=0$), the
whole extended Kalman filter \eqref{eq:KFdefF}-\eqref{eq:KFsupdate}
is equivalent to
\begin{align}
P_t^{-1} &\gets P_{t-1}^{-1}+\transp{H_t} R_t^{-1} H_t
\\
s_t &\gets s_{t-1} - P_t \transp{\left(\frac{\partial \ell_t(y_t)}{\partial
s_{t-1}}\right)}
\end{align}
\end{lem}

\begin{dem}
The first statement is well-known for Kalman filters
\cite[(6.33)]{simon2006kalmanbook}. Indeed, expanding
the definition of $K_t$ in the update \eqref{eq:KFPupdate} of $P_t$, we have 
\begin{equation}
P_t=P_{t|t-1}- P_{t|{t-1}}\transp{H_{t}}\left(H_{t} P_{t|{t-1}}
\transp{H_{t}}+R_{t}\right)^{-1} H_t P_{t|t-1}
\end{equation}
but this is equal to 
$(P_{t|t-1}^{-1}+\transp{H_t} R_t^{-1} H_t)^{-1}$
 thanks to the Woodbury matrix identity.

The second statement follows from Proposition~\ref{prop:Kalmanasgrad} and the
fact that for $f(s,u)=s$, the transition step of the Kalman filter is
just $s_{t|t-1}=s_{t-1}$ and $P_{t|t-1}=P_{t-1}$.
\end{dem}

\begin{lem}
\label{lem:hrh}
For
exponential families $p(y|\hat y)$, the term $\transp{H_t} R_t^{-1} H_t$
appearing in Lemma~\ref{lem:infofilter} is
equal to the Fisher information matrix of $y$ with respect to the state
$s$,
\[
\transp{H_t} R_t^{-1} H_t=\E_{y\sim p(y|\hat y_t)}\left[
\frac{
\partial \ell_t(y)}{\partial s_{t|t-1}}^{\otimes 2}
\right]
\]
where $\ell_t(y)=-\ln p(y|\hat y_t)$ depends on $s$ via $\hat y=h(s,u)$.
\end{lem}

\begin{dem}
Let us omit time indices for brevity. We have $\ds\frac{\partial
\ell(y)}{\partial s}=\frac{\partial \ell(y)}{\partial \hat y}\frac{\partial
\hat y}{\partial s}=\frac{\partial \ell(y)}{\partial \hat y}H$.
Consequently, $\ds \E_y
\left[\frac{\partial \ell(y)}{\partial s}^{\otimes 2}\right]=\transp{H}\,
\E_y
\left[\frac{\partial \ell(y)}{\partial \hat y}^{\otimes 2}\right] H$. 
The middle term $\ds \E_y
\left[\frac{\partial \ell(y)}{\partial \hat y}^{\otimes 2}\right]$ is the
Fisher matrix of the random variable $y$ with respect to $\hat y$.

Now, for
an exponential family $y\sim p(y|\hat y)$ in mean parameterization $\hat y$, the Fisher
matrix with respect to $\hat y$ is equal to the inverse covariance matrix of
the sufficient statistics of $y$ (Appendix, \eqref{eq:fishbarT}), that
is, $R_t^{-1}$.
\end{dem}

\begin{dem}[ of Theorem~\ref{thm:natkal}]
By induction on $t$. By the combination of Lemmas~\ref{lem:infofilter}
and~\ref{lem:hrh}, the update of the Kalman filter with static dynamics
($s_{t|t-1}=s_{t-1}$)
is
\begin{align}
P_t^{-1} &\gets P_{t-1}^{-1}+\E_{y\sim p(y|\hat y_t)}\left[
\frac{
\partial \ell_t(y)}{\partial s_{t-1}}^{\otimes 2}
\right]
\\
s_t &\gets s_{t-1} - P_t \transp{\left(\frac{\partial \ell_t(y_t)}{\partial
s_{t-1}}\right)}
\end{align}

Defining $J_t=P_t^{-1}/(t+1)$, this update is equivalent to
\begin{align*}
J_t &\gets \frac{t}{t+1} J_{t-1}+\frac{1}{t+1} \E_{y\sim p(y|\hat y_t)}\left[
\frac{
\partial \ell_t(y)}{\partial s_{t-1}}^{\otimes 2}
\right]
\\
s_t &\gets s_{t-1} -\frac{1}{t+1} J_t^{-1} \transp{\left(\frac{\partial
\ell_t(y_t)}{\partial
s_{t-1}}\right)}
\end{align*}
Under the identification $s_{t-1}\leftrightarrow \theta_{t-1}$,
this is the online natural gradient update with learning rate $\eta_t=1/(t+1)$
and metric update rate $\gamma_t=1/(t+1)$.
\end{dem}

The proof of Proposition~\ref{prop:rates} is similar, with additional
factors $(1-\lambda_t)$. Proposition~\ref{prop:regul} is proved by
applying a fading memory Kalman filter to a
modified log-likelihood $\bar L_0\deq n_\mathrm{prior}\ln \pi(\theta)$,
$\bar L_t\deq \ln p_\theta(y_t) + (1-\lambda_t)\bar L_{t-1}+\lambda_t
n_\mathrm{prior}
\ln \pi(\theta)$ so that the prior is kept constant in $\bar L_t$.

\section{Natural gradient as a Kalman filter: the state space (recurrent)
case}
\label{sec:rec}

\subsection{Recurrent models, RTRL}

Let us now consider non-memoryless models, i.e., models defined by a
recurrent or state space equation
\begin{equation}
\label{eq:recmodel}
\hat y_t=\transf(\hat y_{t-1},\theta,u_t)
\end{equation}
with $u_t$ the observations at time $t$.
To save notation, here we dump into $\hat y_t$ the whole state of the
model, including both the part that contains the prediction about $y_t$
and all state or internal variables (e.g.,
all internal and output layers of a recurrent neural network, not only
the output layer). The state $\hat y_t$, or a part
thereof, defines a loss function $\ell_t(y_t)\deq -\ln p(y_t|\hat y_t)$ for each
observation $y_t$.

The current state $\hat y_t$ can be seen as a function which
depends on $\theta$ via the whole trajectory. The derivative of the
current state with respect to $\theta$ can be computed inductively just
by differentiating the recurrent equation \eqref{eq:recmodel} defining $\hat y_t$:
\begin{equation}
\frac{\partial \hat y_t}{\partial \theta}=\frac{\partial \transf(\hat
y_{t-1},\theta,u_t)}{\partial \theta}+\frac{\partial \transf(\hat
y_{t-1},\theta,u_t)}{\partial \hat y_{t-1}}\frac{\partial \hat
y_{t-1}}{\partial \theta}
\end{equation}

\emph{Real-time recurrent learning} \cite{Jaeger_tutorial} uses this equation to keep an estimate
$G_t$ of $\frac{\partial \hat y_t}{\partial \theta}$. 
RTRL then
uses $G_t$ to estimate the gradient of the loss function $\ell_t$ with
respect to $\theta$ via the chain rule, $\partial \ell_t/\partial
\theta=(\partial \ell_t/\partial \hat y_t) (\partial \hat y_t/\partial
\theta)=(\partial \ell_t/\partial \hat y_t) G_t$.

\begin{defi}[ (Real-time recurrent learning)]
Given a recurrent model $\hat y_t = \transf(\hat
y_{t-1},\theta_{t-1},u_t)$, real-time recurrent learning (RTRL) learns the
parameter $\theta$ via
\begin{align}
\label{eq:rtrl}
G_t &\gets \frac{\partial
\transf}{\partial \theta_{t-1}}+\frac{\partial \transf}{\partial \hat
y_{t-1}}\,G_{t-1},
\qquad G_0\deq 0
\\g_t &\gets \frac{\partial \ell_t (y_t)}{\partial\hat y_t} \, G_t
\\\theta_t &\gets \theta_{t-1} - \eta_t \transp{g_t}
\end{align}
\end{defi}

Since $\theta$ changes at each step, the actual estimate $G_t$ in RTRL is
only an approximation of the gradient $\frac{\partial \hat y_t}{\partial
\theta}$ at $\theta=\theta_t$, valid in the limit of small learning rates $\eta_t$.

In practice, RTRL has a high computational cost due to the necessary
storage of $G_t$, a matrix of size $\dim\theta\times \dim \hat y$. For
large-dimensional models, backpropagation through time is usually
preferred, truncated to a certain length in the past
\cite{Jaeger_tutorial}; \cite{nobacktrack,uoro} introduce a low-rank,
unbiased approximation of $G_t$.

\subsection{Statement of the correspondence, recurrent case}
\label{sec:rec_statement}

There are several ways in which a Kalman filter can be used to
estimate $\theta$ for such recurrent models.
\begin{enumerate}
\item A first possibility is to view each $\hat y_t$ as a function of $\theta$
via the whole trajectory, and to apply a Kalman filter on $\theta$. This
would require, in principle, recomputing the whole trajectory from time
$0$ to time $t$ using the new value of $\theta$ at each step, and using
RTRL to compute $\partial \hat y_t/\partial \theta$, which is needed in
the filter. In
practice, the past trajectory is not updated, and truncated backpropagation
through time is used to approximate the derivatice $\partial \hat
y_t/\partial \theta$ \cite{Jaeger_tutorial, Haykin_book}.
\item A second possibility is the \emph{joint Kalman filter},
namely, a Kalman filter on the pair
$(\theta,\hat y_t)$ \cite[\S5]{Haykin_book},
\cite[\S13.4]{simon2006kalmanbook}.
This does not require going back in time, as $\hat
y_{t}$ is a function of $\hat y_{t-1}$ and $\theta$. This 
is the version appearing in
Theorem~\ref{thm:rtrlkal} below. 
\item A third possibility is the \emph{dual Kalman filter}
\cite{WN96dual}: a
Kalman filter for $\theta$ given $\hat y$, and another one for $\hat y$
given $\theta$. This requires to explicitly couple the two Kalman filters
by manually adding RTRL-like terms to account for the (linearized) dependency of
$\hat y$ on $\theta$ \cite[\S5]{Haykin_book}.
\end{enumerate}

Intuitively, the joint Kalman filter maintains a covariance matrix on
$(\theta,\hat y_t)$, whose
off-diagonal term is the covariance between $\hat y_t$ and $\theta$. This
term
captures how the current state would change if another value of the
parameter had been used. The decomposition \eqref{eq:kalmancov} in the
theorem makes this intuition precise in
relation to RTRL: the Kalman covariance between $\hat y_t$ and $\theta$ is
directly given by the RTRL gradient $G_t$.

\begin{thm}[ (Kalman filter on $(\theta,\hat y)$ as RTRL+natural
gradient+state correction)]%
\label{thm:rtrlkal}%
Consider a recurrent model $\hat y_t = \transf(\hat
y_{t-1},\theta_{t-1},u_t)$. Assume that the observations $y_t$ are
predicted with
a probabilistic model $p(y|\hat y_t)$ that is an exponential family
with mean parameter a subset of $\hat y_t$.

Given an estimate $G_t$ of $\partial \hat y_t/\partial \theta$, and an
observation $y$, denote
\begin{equation}
\label{eq:gradlossestim}
g_t(y)\deq \frac{\partial \ell_t(y)}{\partial\hat y_t} \, G_t
\end{equation}
the corresponding estimate of $\partial \ell_t(y)/\partial \theta$.

Then these two algorithms are equivalent:
\begin{itemize}
\item The
extended Kalman filter on the pair $(\theta,\hat
y)$ with transition function $(\Id,\transf)$, initialized with covariance matrix $P_0^{(\theta,\hat y)}=\begin{pmatrix} \Ptheta_0 &
0 \\ 0 &0 \end{pmatrix}$, and with no process noise ($Q=0$). 
\item A natural gradient RTRL algorithm with learning rate
$\eta_t=1/(t+1)$, defined as follows.
The state, RTRL gradient and Fisher matrix have a transition step
\begin{align}
\hat y_t &\gets\transf(\hat y_{t-1},\theta_{t-1},u_t)
\\
G_t &\gets \frac{\partial
\transf}{\partial \theta_{t-1}}+\frac{\partial \transf}{\partial \hat
y_{t-1}}\,G_{t-1},
\qquad G_0\deq 0
\\
J_t &\gets (1-\eta_t) J_{t-1} + \eta_t \E_{y\sim p(y|\hat y_t)}
\left[g_t(y)^{\otimes 2}\right],
\qquad J_0\deq (\Ptheta_0)^{-1} \label{eq:recJupdate}
\end{align}
and after observing $y_t$,
the state and
parameter are updated as
\begin{align}
\deltatheta &\gets J_t^{-1}\transp{g_t(y_t)}
\\\theta_t &\gets \theta_{t-1}-\eta_t \,\deltatheta
\\\hat y_t &\gets \hat y_t-\eta_t \, G_t \,\deltatheta
\label{eq:jointstateupdate}
\end{align}
\end{itemize}
Moreover, at each time $t$, the covariance matrix of the
extended Kalman filter over $(\theta,\hat y)$ is related to $G_t$ and
$J_t$ via
\begin{equation}
\label{eq:kalmancov}
P_t^{(\theta,\hat y)}=\eta_t
\begin{pmatrix}
J_t^{-1} & J_t^{-1} \transp{G_t}
\\ G_t J_t^{-1} & G_t J_t^{-1} \transp{G_t}
\end{pmatrix}
\end{equation}
\end{thm}

This result may explain an observation from
\cite[\S4.2]{williams1992training} that
RTRL can be obtained by introducing some drastic simplifications
in the Kalman filter equations (changing the formula of the Kalman
optimal gain and neglecting the covariance matrix update).


Again,
the expectation for the Fisher matrix in \eqref{eq:recJupdate}
may be estimated by a Monte Carlo sample $y\sim p(y|\hat y_t)$, or by
just using the current observation $y=y_t$, as discussed
after Definition~\ref{def:natgrad}.

As before, learning rates $\eta_t$ different from $1/(t+1)$ can be
obtained by introducing a fading memory (i.e., process noise $Q$
proportional to $P$) in the joint Kalman filter. We omit the statement for simplicity, but it is
analogous to Propositions~\ref{prop:rates} and \ref{prop:regul}.

The algorithm above features a state update \eqref{eq:jointstateupdate}
together with the parameter update; this is not commonly used in online
recurrent neural network algorithms. In small-scale experiments, we have
not found any clear effect of this; besides, such state updates must
be applied cautiously if the range of possible values for the state $\hat y$ is
somehow constrained.

In the result above, the Kalman filter is initialized with a
covariance matrix in which every uncertainty comes from uncertainty on
$\theta$ rather than the initial state $\hat y_0$. This has the advantage
of making the correspondence algebraically simple, but is not a
fundamental restriction. 
If modelling an initial uncertainty on $\hat y_0$ is important, one can
always apply the theorem by incorporating the initial condition as an additional component of the
parameter $\theta$, with its own variance; in this case, 
$G_0$ must be
initialized to $\Id$ on the corresponding component of $\theta$, namely
\begin{equation}
\theta^{+\mathrm{init}}\deq \transp{(\theta,\hat y_0)},\qquad
G_0\deq \frac{\partial \hat y_0}{\partial \theta^{+\mathrm{init}}}=(0,\Id)
\end{equation}
and then Theorem~\ref{thm:rtrlkal} can be applied to
$\theta^{+\mathrm{init}}$.

Actually this operation is often not needed at all: indeed, if the dynamical
system is such that the initial condition is forgotten reasonably
quickly, then the initial covariance of $\hat y_0$ decreases (terms $W$
in the proof below) and the Kalman covariance tends to the type
\eqref{eq:kalmancov} above exponentially fast, even without using
$\theta^{+\mathrm{init}}$. This is the case, for
instance, for any stable linear dynamical system, as a consequence of
Lemmas \ref{lem:transdecomp}-\ref{lem:obsdecomp}, and more generally for
any system with geometric memory in the sense that $\frac{\partial \hat
y_{t}}{\partial \hat y_{t-1}}$ is contracting for a fixed parameter and a
given input.

The filtering literature contains updates similar to the above for $G_t$, but more
complex \cite{ljung83,Haykin_book}; this is, first, because they are
expressed over the variable $\Cov(\hat y_t,\theta)=G_tJ_t^{-1}$ instead
of $G_t$ alone, second, because we have initialized the uncertainty on
$\hat y_0$ to $0$, and, third, because in dual rather than joint filter
approaches, higher-order terms depending on second derivatives of $F$ are
sometimes included.  Interestingly, there is some debate in the
literature about whether to add some second-order corrections to the joint Kalman
filter (especially \cite[\S2.3.3]{ljung83}, see discussion in
\cite[\S5.3.4]{Haykin_book}). The interpretation in
Theorem~\ref{thm:rtrlkal}
makes it clear which terms are neglected:
in particular, in RTRL $G_t$ is not recomputed after the update of $\theta$ and
$\hat y_t$, so that
$G_t$ contains a mixture of derivatives at different values of $\theta$
over time. Correcting for this would involve second derivatives of $F$ (as
in \cite[\S5, Appendix A]{Haykin_book}), thus amounting to a partial
implementation of a second-order extended Kalman filter (EKF2,
\cite[\S13.3]{simon2006kalmanbook}).

In terms of computational cost, for recurrent neural networks (RNNs),
RTRL alone is already as costly as the joint Kalman filter \cite{williams1992training}.  Indeed, RTRL requires $(\dim \theta)$ forward tangent
propagations at each step, each of which costs $O(\dim \theta)$ for a
standard RNN model \cite{Jaeger_tutorial}, thus for a total cost of
$O((\dim\theta)^2)$ per time step. The Fisher matrix is of size
$(\dim\theta)^2$; if a single Monte Carlo sample $y\sim p(y|\hat y_t)$ is
used, then the Fisher matrix update is rank-one and costs
$O((\dim\theta)^2)$; the update of the \emph{inverse} Fisher matrix can
be maintained at the same cost thanks to the Woodbury matrix identity
(as done, e.g., in \cite{TONGA}). Thus, if RTRL is computationally
affordable, there is little point in not using the Fisher matrix on top.

\subsection{Proofs for the recurrent case}

We now turn to the proof for the recurrent case, involving a joint Kalman
filter on $(\theta,\hat y)$. The key is to decompose the Kalman
covariance matrix of the pair $(\theta,\hat y)$ into three variables
\eqref{eq:Pdecomp}: the covariance of $\theta$, the correlation between
$\theta$ and $\hat y$, and the part of the covariance of $\hat y$ that
does not come from its correlation with $\theta$ (its so-called Schur
complement). This provides a nice expression for the transition step of
the Kalman filter (Lemma~\ref{lem:transdecomp}).

Then we find that the correlation between $\theta$ and $\hat y$ is
exactly the gradient $G=\frac{\partial\hat y}{\partial \theta}$
maintained by RTRL (Corollary~\ref{cor:kaldecomp}); meanwhile, we find $\theta$
and its covariance essentially follow a standalone Kalman filter related
to the observations via $G$, which is a natural gradient for the same
reasons as in the static case.

In the recurrent case,
we are applying an extended Kalman filter to the state $s=\begin{pmatrix}\theta\\\hat
y\end{pmatrix}$ with
transition function $f=\begin{pmatrix}\Id \\ \transf\end{pmatrix}$.
Let us decompose the covariance matrix $P_t$ of this system as
\begin{equation}
P_t=
\begin{pmatrix}
\Ptheta_t & \transp{(\Pthetay_t)}
\\ \Pthetay_t & \Py_t
\end{pmatrix}
\end{equation}

From now on, for simplicity we omit the time indices when no ambiguity is
present.

By the theory of Schur complement for positive-semidefinite matrices
\cite[Appendix A.5.5]{BVconvopt},
letting $P^+$ be any generalized inverse of $\Ptheta$, we know that
$\Py-\Pthetay P^+\transp{{\Pthetay}}$ is positive-semidefinite and that
$\Pthetay (\Id-P^+\Ptheta)=0$. The latter rewrites as $\Pthetay=\Pthetay
P^+ \Ptheta$. Let us set 
\begin{equation}
W\deq \Py-\Pthetay
P^+\transp{{\Pthetay}}, \qquad G\deq \Pthetay P^+
\end{equation}
Then
$\Pthetay=G\Ptheta$ and $W=\Py-G\Ptheta \transp{G}$. Thus, at each time $t$ we can decompose $P_t$
as
\begin{equation}
\label{eq:Pdecomp}
P_t=
\begin{pmatrix}
\Ptheta & \transp{(G\Ptheta)}
\\ G\Ptheta & W+G\Ptheta \transp{G}
\end{pmatrix}
\end{equation}
without loss of generality, where $W$ is positive-semidefinite. This decomposition tells us which part of the
covariance of the current state $\hat y$ comes from the covariance of the
parameter $\theta$ via the dynamics of the system.

First, we will show that if $W_0=0$, then $W_t=0$ for all $t$, and that
in this case $G_t$ satisfies the RTRL equation.

\begin{lem}
\label{lem:transdecomp}
Consider the extended Kalman filter on the pair $s=\transp{(\theta, \hat
y)}$ with
transition function $f=\transp{(\theta,\transf(\hat y,\theta,u))}$ and no
added noise ($Q_t=0$). Then the Kalman transition step
\eqref{eq:transP} on $P$, expressed in the decomposition
\eqref{eq:Pdecomp}, is
equivalent to
\begin{align}
\Ptheta &\gets \Ptheta
\\
W&\gets \frac{\partial \transf}{\partial \hat
y}W\transp{\frac{\partial \transf}{\partial \hat y}}
\label{eq:transW}
\\
G&\gets \frac{\partial \transf}{\partial \theta}+\frac{\partial
\transf}{\partial \hat y}G
\end{align}
\end{lem}

This equation for $G$ is the RTRL update.

\begin{dem}[ of the lemma]
This is a direct computation using the Kalman transition step
\eqref{eq:transP} for $P$.
Indeed, the decomposition \eqref{eq:Pdecomp} of $P$ rewrites as
\begin{equation}
\label{eq:Pdecompprod}
P_t=
\begin{pmatrix} \Id & 0\\G &\Id \end{pmatrix}
\begin{pmatrix} \Ptheta &0\\0 &W \end{pmatrix}
\begin{pmatrix} \Id & \transp{G}\\0& \Id \end{pmatrix}
\end{equation}
Now, the Kalman transition step \eqref{eq:transP} for $P$ is
$P_{t|t-1}=\frac{\partial f}{\partial s}P_{t-1}
\transp{\frac{\partial f}{\partial s}}$ when $Q=0$.
So the update is equivalent to replacing $\begin{pmatrix} \Id & 0\\G &\Id
\end{pmatrix}$ with $\frac{\partial f}{\partial s}
\begin{pmatrix} \Id & 0\\G &\Id
\end{pmatrix}$ on both sides in \eqref{eq:Pdecompprod}.
Here we have $\frac{\partial
f}{\partial s}=\begin{pmatrix}
\Id & 0 \\
\frac{\partial \transf}{\partial \theta} & \frac{\partial \transf}{\partial
\hat y}
\end{pmatrix}$.
This yields the result.
\end{dem}

\begin{lem}
\label{lem:obsdecomp}
Consider the extended Kalman filter on the pair $s=\transp{(\theta, \hat
y)}$ with
transition function $f=\transp{(\theta,\transf(\hat y,\theta,u))}$. Then
the observation update
\eqref{eq:KFdefK}--\eqref{eq:KFPupdate} of $P_t$, expressed in the
variables $\Ptheta$, $W$, and $G$,
is given by
\begin{align}
\label{eq:recPupdate}
\Ptheta&\gets
\Ptheta-\Ptheta\transp{G}(W+R+G\Ptheta\transp{G})^{-1}G\Ptheta
\\
\label{eq:recWupdate}
W&\gets W-W(W+R)^{-1}W
\\
G&\gets (\Id -R^{-1}W) G
\end{align}
in that order, where $R$ is given by \eqref{eq:KFdefR}. Moreover, if
$\Ptheta$ or $W$ are invertible then their respective updates are
equivalent to
\begin{equation}
\label{eq:recPinvupdate}
(\Ptheta)^{-1}\gets (\Ptheta)^{-1}+\transp{G}(W+R)^{-1}G
\end{equation}
and
\begin{equation}
\label{eq:recWinvupdate}
W^{-1}\gets W^{-1}+R^{-1}
\end{equation}
\end{lem}

Thus, the updates for $W$, \eqref{eq:transW} and \eqref{eq:recWupdate}, 
are just the updates of an extended Kalman
filter on $\hat y$ alone, with covariance matrix $W$ and noise measurement $R$. The update for
$\Ptheta$ is identical to an extended Kalman filter on $\theta$ where
measurements are made on $\hat y$, with $\hat y$ 
seen as a function of $\theta$ with derivative $\partial \hat y/\partial
\theta=G$, and where the measurement noise on $\hat y$ is $R+W$ (the
measurement noise on $y$ plus the covariance of $\hat y$). Thus, these two lemmas relate the joint Kalman filter on
$(\theta,\hat y)$ to the dual Kalman filter that filters separately
$\theta$ given $\hat y$ and $\hat y$ given $\theta$, together with an
estimate of $\partial \hat y/\partial \theta$. As far as we could check,
this decomposition
is specific to a situation in which one component (the parameter) is
supposed to have static underlying dynamics, $\theta_{t+1}=\theta_t$.

\begin{dem}[ of the lemma]
In our case, the function $h$ of the extended Kalman filter is the
function that sends $(\theta,\hat y)$ to $\hat y$. In particular,
$H_t=(0,\Id)$.

First, if $\Ptheta$ and $W$ are invertible, then the updates
\eqref{eq:recPupdate}, \eqref{eq:recWupdate} for
$\Ptheta$ and $W$ follow from the updates \eqref{eq:recPinvupdate},
\eqref{eq:recWinvupdate} on their inverses, thanks to
the Woodbury matrix identity. Since working on the inverses is simpler,
we shall prove only the latter. Since \eqref{eq:recPupdate},
\eqref{eq:recWupdate} are continuous in $\Ptheta$ and $W$, the
non-invertible case follows by continuity.

Starting again with the decomposition of $P_t$ as a product
\eqref{eq:Pdecompprod}, the inverse of $P_t$ is
\begin{align}
P_t^{-1}&=
\begin{pmatrix} \Id & \transp{G}\\0& \Id \end{pmatrix}^{-1}
\begin{pmatrix} \Ptheta &0\\0 &W \end{pmatrix}^{-1}
\begin{pmatrix} \Id & 0\\G &\Id \end{pmatrix}^{-1}
\\&=
\begin{pmatrix}
(\Ptheta)^{-1}+\transp{G}W^{-1}G & -\transp{G} W^{-1}
\\
-W^{-1}G & W^{-1}
\end{pmatrix}
\label{eq:Pinvdecomp}
\end{align}

From Lemma~\ref{lem:infofilter}, the Kalman observation udpate for $P_t$
amounts to adding $\transp{H}R^{-1}H$ to $P_t^{-1}$. Here
$H=(0,\Id)$ so that $\transp{H}R^{-1}H$ is $
\begin{pmatrix}
0 & 0 \\0 & R^{-1}
\end{pmatrix}$. So the update for $P_t$ amounts to
\begin{equation}
\label{eq:Pinvnewdecomp}
P_t^{-1}\gets 
\begin{pmatrix}
(\Ptheta)^{-1}+\transp{G}W^{-1}G & -\transp{G} W^{-1}
\\
-W^{-1}G & W^{-1} + R^{-1}
\end{pmatrix}
\end{equation}

To interpret this as an update on $\Ptheta$, $W$ and $G$, we have to
introduce new variables $\tilde W$, $\tilde G$, and $\tilde \Ptheta$ such
that \eqref{eq:Pinvnewdecomp} takes the original form
\eqref{eq:Pinvdecomp} in these new variables.

Introducing $\tilde W^{-1}\deq W^{-1}+R^{-1}$, 
the update rewrites as
\begin{equation}
P_t^{-1}\gets 
\begin{pmatrix}
(\Ptheta)^{-1}+\transp{G}W^{-1}G & -\transp{G} (\Id -R^{-1}\tilde
W)\tilde W^{-1}
\\
-\tilde W^{-1}(\Id - \tilde WR^{-1})G & \tilde W^{-1}
\end{pmatrix}
\end{equation}
Introducing $\tilde G\deq (\Id-\tilde WR^{-1})G$ and $(\tilde
\Ptheta)^{-1}\deq (\Ptheta)^{-1}+\transp{G}W^{-1}G-\transp{\tilde
G}\tilde W^{-1}\tilde G$, we get back the original form
\eqref{eq:Pinvdecomp}. This provides the updates $\tilde W$ and
$\tilde G$ for $W$ and $G$. We
still have to find a more explicit expression for $(\tilde
\Ptheta)^{-1}$.

Since we have defined $\tilde W$ and $\tilde G$ by identifying
\eqref{eq:Pinvnewdecomp} with the original form \eqref{eq:Pinvdecomp}, we
have $\tilde W\tilde G=WG$ by construction. Thus
\begin{align}
(\tilde
\Ptheta)^{-1}&= (\Ptheta)^{-1}+\transp{G}W^{-1}G-\transp{\tilde
G}\tilde W^{-1}\tilde G
\\&=(\Ptheta)^{-1}+\transp{G}W^{-1}G-\transp{\tilde
G}\tilde W^{-1}\tilde W\tilde W^{-1}\tilde G
\\&=(\Ptheta)^{-1}+\transp{G}W^{-1}G-\transp{G}W^{-1}\tilde W
W^{-1}G
\end{align}
Thanks to the identity
$A^{-1}-A^{-1}BA^{-1}=(A+(B^{-1}-A^{-1})^{-1})^{-1}$ for any matrices $A$
and $B$ (this follows from the matrix inversion formula
$(A+C)^{-1}=A^{-1}-A^{-1}(A^{-1}+C^{-1})^{-1}A^{-1}$ applied to
$C=(B^{-1}-A^{-1})^{-1}$), we find
\begin{equation}
W^{-1}-W^{-1}\tilde W W^{-1}=(W+(\tilde W^{-1}-W^{-1})^{-1})^{-1}
\end{equation}
but by definition, $\tilde W^{-1}=W^{-1}+R^{-1}$ so that
\begin{equation}
W^{-1}-W^{-1}\tilde W W^{-1}=(W+R)^{-1}
\end{equation}
thus
\begin{equation}
(\tilde
\Ptheta)^{-1}=(\Ptheta)^{-1}+\transp{G}(W+R)^{-1}G
\end{equation}
which concludes the proof of the lemma.
\end{dem}

Putting the last two lemmas side by side in the case $W=0$, we obtain a
much simpler update.

\begin{cor}
\label{cor:kaldecomp}
Consider the extended Kalman filter on the pair $s=\transp{(\theta, \hat
y)}$ with
transition function $f(s)=\transp{(\theta,\transf(\hat y,\theta,u))}$ and no
added noise ($Q_t=0$). Decompose the covariance $P$ of the state $s$ as in
\eqref{eq:Pdecomp} using $\Ptheta$, $G$, $W$.
If $W=0$ and $\Ptheta$ is invertible then
performing the Kalman transition update followed by the observation update
is equivalent to
\begin{align}
G&\gets \frac{\partial \transf}{\partial \theta}+\frac{\partial
\transf}{\partial \hat y}G
\\
\label{eq:recPinvupdatesimple}
(\Ptheta)^{-1}&\gets (\Ptheta)^{-1}+\transp{G}R^{-1}G
\\
W&\gets 0
\end{align}
in that order.
\end{cor}

From this, the end of the proof of Theorem~\ref{thm:rtrlkal} essentially proceeds
as in the non-recurrent case. Since we initialize $W$ to $0$ in
Theorem~\ref{thm:rtrlkal}, we have
$W=0$ at all times. As before, for exponential families
$R^{-1}$ is equal to the Fisher matrix with respect to $\hat y_t$,
namely, $R^{-1}=\E_{y \sim p(y|\hat y)}\ds
\left[\frac{\partial \ell_t(y)}{\partial \hat y_t}^{\otimes 2}\right]$ 
(Appendix).
Now, the term $\E_{y \sim p(y|\hat y)}
\left[g_t(y)^{\otimes
2}\right]$ in the Fisher matrix update \eqref{eq:recJupdate} uses
$g_t(y)=\ds\frac{\partial \ell_t(y)}{\partial\hat y_t} \, G_t$
\eqref{eq:gradlossestim} to estimate the derivative of the loss
$\ell_t(y)$
with respect to $\theta$. So
the term $\transp{G}R^{-1}G$ in \eqref{eq:recPinvupdatesimple} coincides with the Fisher
matrix update term $\E_{y \sim p(y|\hat y)} \left[g_t(y)^{\otimes
2}\right]$ in \eqref{eq:recJupdate}. (Compare Lemma~\ref{lem:hrh}.)
So if we just define $J_t\deq \eta_t (\Ptheta_t)^{-1}$ with
$\eta_t=1/(t+1)$, the additive update \eqref{eq:recPinvupdatesimple} on
$(\Ptheta)^{-1}$
translates as the online Fisher matrix update \eqref{eq:recJupdate} on
$J_t$.

Moreover, since the Kalman gradient is an ordinary gradient
preconditioned with the covariance matrix $P_t$
(Proposition~\ref{prop:Kalmanasgrad}), the update of the pair
$(\theta,\hat y)$ is
\begin{equation}
\begin{pmatrix} \theta_t \\ \hat y_t \end{pmatrix}
\gets \begin{pmatrix} \theta_{t-1} \\ \hat y_{t} \end{pmatrix} 
-P_t \begin{pmatrix} 0 \\ \ds\transp{\frac{\partial \ell_t}{\partial \hat
y_t}}\end{pmatrix}
\end{equation}
(indeed $\ell_t$ does not depend explicitly on $\theta$ in recurrent
models, only via the current state $\hat y_t$).
Given the decomposition $P_t=
\begin{pmatrix}
\Ptheta & \transp{(G \Ptheta)}
\\ G \Ptheta & G \Ptheta \transp{G}
\end{pmatrix}
$, this translates as
\begin{equation}
\begin{pmatrix} \theta_t \\ \hat y_t \end{pmatrix}
\gets \begin{pmatrix} \theta_{t-1} \\ \hat y_{t} \end{pmatrix} 
-\begin{pmatrix} \Ptheta  \\ G \Ptheta \end{pmatrix}\transp{\left(\frac{\partial \ell_t}{\partial \hat
y_t}G\right)}
\end{equation}
which is the update in Theorem~\ref{thm:rtrlkal}.


{\small

\appendix

\section{Appendix: reminder on exponential families}

\newcommand{\tsum}{{\textstyle \sum}}

An \emph{exponential family of probability distributions} on a variable
$x$ (discrete or continuous), with \emph{sufficient statistics}
$T_1(x),\ldots,T_K(x)$, is the
following family of distributions, parameterized by $\beta\in \R^K$:
\begin{equation}
p_\beta(x)=\frac{1}{Z(\beta)}\,\mathrm{e}^{\sum_k \beta_k T_k(x)}\,\lambda(\d x)
\end{equation}
where $Z(\beta)$ is a normalizing constant, and $\lambda(\d x)$ is any reference
measure on $x$, such as the Lebesgue measure or any discrete measure. The
family is obtained by varying the parameter $\beta\in \R^K$, called the
\emph{natural} or \emph{canonical} parameter. We will assume that the
$T_k$ are linearly independent as functions of $x$ (and linearly
independent from the constant function); this ensures that
different values of $\beta$ yield distinct distributions.

For instance, Bernoulli distributions are obtained with $\lambda$
the uniform measure on $x\in \{0,1\}$ and with a single sufficient
statistic $T(0)=0$, $T(1)=1$. Gaussian
distributions with a fixed variance are obtained with $\lambda(\d
x)$ the Gaussian distribution centered on $0$, and $T(x)=x$.

Another, often convenient parameterization of the same family is the
following: each value of $\beta$ gives rise to an average value $\bar T$
of the sufficient statistics, \begin{equation} \bar T_k\deq \E_{x\sim
p_\beta} T_k(x) \end{equation} For instance, for Gaussian distributions
with fixed variance, this is the mean, and for a Bernoulli variable this
is the probability to sample $1$.

Exponential families satisfy the identities
\begin{equation}
\label{eq:expder}
\frac{\partial \ln p_\beta(x)}{\partial \beta_k}=T_k(x)-\bar T_k,\qquad
\frac{\partial \ln Z}{\partial \beta_k}=\bar T_k
\end{equation}
by a simple computation \cite[(2.33)]{Amari2000book}.

These identities are useful to compute the Fisher matrix $J_\beta$ with
respect to the variable $\beta$, as follows \cite[(3.59)]{Amari2000book}:
\begin{align}
(J_\beta)_{ij} &\deq \E_{x\sim p_\beta} \left[
\frac{\partial \ln p_\beta(x)}{\partial \beta_i}
\frac{\partial \ln p_\beta(x)}{\partial \beta_j}
\right]
\\&= \E_{x\sim p_\beta} \left[ (T_i(x)-\bar T_i)(T_j(x)-\bar T_j)\right]
\\&= \Cov (T_i,T_j)
\label{eq:J=cov_comp}
\end{align}
or more synthetically
\begin{equation}
J_\beta=\Cov(T)
\end{equation}
where the covariance is under the law $p_\beta$. That is, for exponential
families the Fisher matrix
is the covariance matrix of the sufficient statistics. In particular it
can be estimated empirically, and is sometimes known algebraically.

In this work we need the Fisher matrix with respect to the mean parameter
$\bar T$,
\begin{equation}
(J_{\bar T})_{ij}=\E_{x\sim p_\beta}\left[
\frac{\partial \ln p_\beta(x)}{\partial \bar T_i}
\frac{\partial \ln p_\beta(x)}{\partial \bar T_j}
\right]
\end{equation}
By substituting $\frac{\partial \ln p(x)}{\partial \alpha}=\frac{\partial \ln
p(x)}{\partial \beta}\frac{\partial \beta}{\partial \alpha}$,
the Fisher matrices
$J_\alpha$ and $J_\beta$ with respect to parameterizations $\alpha$ and $\beta$ are
related to each other via
\begin{equation}
J_\alpha=\transp{\frac{\partial \beta}{\partial \alpha}}J_\beta\, \frac{\partial \beta}{\partial
\alpha}
\end{equation}
(consistently with the interpretation of the Fisher matrix as a
Riemannian metric and the behavior of metrics under change of coordinates
\cite[\S 2.3]{GHL87}).
So we need to compute $\partial{\bar T}/\partial \beta$. Using the
log-trick \begin{equation}
\partial \E_{x\sim p} f(x)=\E_{x\sim p} \left[f(x)\,\partial \ln
p(x)\right]
\end{equation}
together with \eqref{eq:expder}, we find
\begin{align}
\label{eq:expjac}
\frac{\partial \bar T_i}{\partial \beta_j}=
\frac{\partial \E T_i(x)}{\partial \beta_j}
=\E\left[ T_i(x)(T_j(x)-\bar T_j)
\right]=\E\left[ (T_i(x)-\bar T_i)(T_j(x)-\bar T_j)
\right]=(J_\beta)_{ij}
\end{align}
so that 
\begin{equation}
\frac{\partial \bar T}{\partial \beta}=J_\beta
\end{equation}
(see \cite[(3.32)]{Amari2000book},
where $\eta$ denotes the mean parameter) and consequently
\begin{equation}
\frac{\partial \beta}{\partial \bar T}=J_\beta^{-1}
\end{equation}
so that we find the Fisher matrix with respect to $\bar T$ to be
\begin{align}
J_{\bar T}&=\transp{\frac{\partial \beta}{\partial \bar T}}J_\beta\, \frac{\partial \beta}{\partial \bar T}
\\&=J_\beta^{-1} J_\beta J_\beta^{-1}
\\&=J_\beta^{-1}=\Cov(T)^{-1}
\label{eq:fishbarT}
\end{align}
that is, the Fisher matrix with respect to $\bar T$ is the inverse
covariance matrix of the sufficient statistics.

This gives rise to a simple formula for the natural gradient of
expectations with respect to the mean parameters. Denoting $\tilde
\nabla$ the natural gradient,
\begin{align}
\tilde\nabla_{\bar T}\, \E f(x) &\deq J_{\bar T}^{-1} \transp{\frac{\partial \E
f(x)}{\partial \bar T}}
\\&= J_{\bar T}^{-1} \,\transp{\frac{\partial \beta}{\partial \bar T}} \,\transp{\frac{\partial \E
f(x)}{\partial \beta}}
\\&=J_{\beta} J_\beta^{-1} \,\transp{\frac{\partial \E
f(x)}{\partial \beta}}
\\&=\transp{\frac{\partial \E
f(x)}{\partial \beta}}
\\&=\E\left[f(x) \,\frac{\partial \ln p_\beta(x)}{\partial \beta}
\right]
\\&=\E\left[f(x) (T(x)-\bar T)
\right]
\\&=\Cov(f,T)
\end{align}
which in particular, can be estimated empirically.
}

\bibliographystyle{alpha}
\bibliography{natkal}

\end{document}